\documentclass[conference]{IEEEtran}
\IEEEoverridecommandlockouts
\usepackage{cite}
\usepackage{amsmath,amssymb,amsfonts}
\usepackage{algorithmic}
\usepackage{graphicx}
\usepackage{textcomp}
\usepackage{xcolor}
\usepackage{subcaption}
\usepackage{caption}

\def\BibTeX{{\rm B\kern-.05em{\sc i\kern-.025em b}\kern-.08em
    T\kern-.1667em\lower.7ex\hbox{E}\kern-.125emX}}

\usepackage{longtable}
\usepackage{tabularx}
\usepackage{xltabular}
\usepackage{footnote}
\makesavenoteenv{tabular}
\usepackage{url}
\usepackage{multicol}
\usepackage{svg}
\usepackage{etoolbox}

\begin{document}
\title{Medical Vision Language Pretraining: A survey
}
\author{Prashant Shrestha$\IEEEauthorrefmark{1}^1$, Sanskar Amgain$\IEEEauthorrefmark{1}^1$, Bidur Khanal$^2$, Cristian A. Linte$^2$, Binod Bhattarai$^3$\\~\\
$^1$~NepAl Applied Mathematics and Informatics Institute for research (NAAMII), Nepal\\
$^2$~Rochester Institute of Technology, Rochester, NY, USA \\
$^3$~School of Natural and Computing Sciences, University of Aberdeen, Aberdeen, UK\\

}

\maketitle
\begingroup\renewcommand\thefootnote{\IEEEauthorrefmark{1}}
\footnotetext{Equal contribution}
\endgroup
\begin{abstract}

Medical Vision Language Pretraining~(VLP) has recently emerged as a promising solution to the scarcity of labeled data in the medical domain. By leveraging paired/unpaired vision and text datasets through self-supervised learning, models can be trained to acquire vast knowledge and learn robust feature representations. Such pretrained models have the potential to enhance multiple downstream medical tasks simultaneously, reducing the dependency on labeled data. However, despite recent progress and its potential, there is no such comprehensive survey paper that has explored the various aspects and advancements in medical VLP. In this paper, we specifically review existing works through the lens of different pretraining objectives, architectures, downstream evaluation tasks, and datasets utilized for pretraining and downstream tasks. Subsequently, we delve into current challenges in medical VLP, discussing existing and potential solutions, and conclude by highlighting future directions. To the best of our knowledge, this is the first survey focused on medical VLP.
\end{abstract}
\vspace{0.5em}

\section{Introduction}
Data-driven artificial intelligence (AI) has undergone rapid advancement in recent years, bringing transformative changes to various domains, including computer vision and natural language processing~\cite{krizhevsky2012imagenet, he2016deep, chen2020simple, vaswani2017attention, devlin2018bert}. The availability of large-scale data has played a pivotal role in driving this progress. With increasing scale, the data has become more multimodal, spanning modalities such as images, text, and other sensor readings. AI is no longer confined to single-modality systems; instead, there has been a notable shift towards multimodal learning~\cite{xu2015show, li2019visualbert, lu2019vilbert, radford2021learning}. Similar trends are quickly emerging, even within the medical domain~\cite{wang2021transbts, dalmaz2022resvit, braman2021deep, zhang2022mmformer}.
\begin{figure} [h!]
    \centering
    \includegraphics[width=0.48\textwidth]{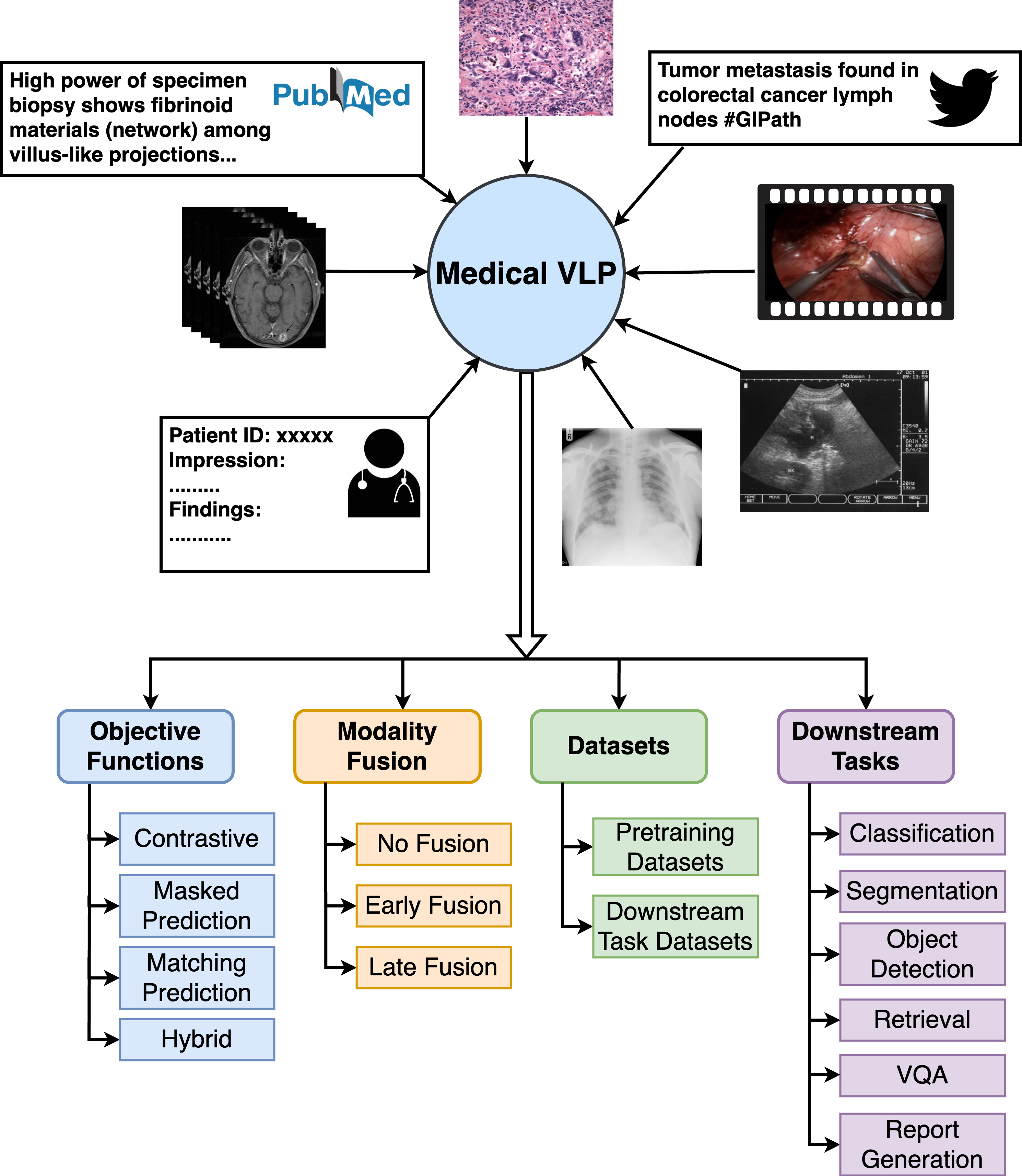}
    \caption{Various aspects of Medical Vision Language Pretraining~(VLP) discussed in this paper.}
    \label{fig:paper_structure}
\end{figure}

Medical data naturally manifests as multimodal information. Often, medical experts rely on information from multiple modalities for diagnostic decision-making. For instance, physicians consider various factors, including medical images, blood test results, and sensor data, to recommend treatments. While past limitations in data availability posed challenges in training expert AI models for the medical domain, several institutions have joined hands to make medical datasets publicly accessible \cite{sudlow2015uk, johnson2019mimic, ikezogwo2023quilt}. Comprising extensive medical images complemented by detailed reports serving as metadata, these multimodal datasets can play a crucial role in training large-scale, generalized AI models.

In recent years, self-supervised learning has become a prominent technique for training with large-scale multimodal medical data \cite{taleb2022contig, taleb2021multimodal, hervella2018retinal}. There is a particular emphasis on vision-language models in both the general domain \cite{radford2021learning, jia2021scaling, geng2022multimodal} and the medical domain \cite{huang2021gloria, wang2022medclip, khare2021mmbert,zhang2022contrastive,dawidowicz2023limitr,wang2022multi}, given that vision and language are two key data modalities. By employing self-supervised techniques, the model can effectively capture extensive knowledge and learn robust representations. Self-supervised learning is often introduced as a pretraining phase, which is later adapted to downstream medical tasks such as classification, segmentation, report generation, etc. without requiring a large dataset to adapt. Medical Vision Language Pretraining (VLP) is a relatively new field that has garnered attention; however, there is a lack of a comprehensive survey paper examining various aspects of recent VLP methods. While \cite{sun2023scoping} explored learning with biomedical images and text, it mostly focused on evaluation tasks and lacked technical depth and a comprehensive overview of pretraining methods—a key aspect we addressed in our survey.

\section{Scope of the Survey}

In this survey, we have covered various deep learning-based medical vision-language pretraining approaches proposed in recent years. We discuss them through the lens of learning objectives, model architectures, medical datasets, and their applications in medical downstream tasks. First, we introduce and mathematically formulate multimodal learning in Section~\ref{multimodal learning} and \ref{multimodal learning approaches}. Subsequently, we delve into the learning objectives and architectures employed by various methods in Section~\ref{Medical VLP Methods} and \ref{architecture}, respectively. Following this, we revisit commonly utilized medical datasets for both VLP and the downstream tasks in Section~\ref{datsets}. Next, we explore different approaches used by various medical downstream tasks to leverage VLP in Section~\ref{downstream evaluation tasks}. Finally, we discuss the limitations and challenges of present VLP methods in Section~\ref{limitations} and the future directions in Section \ref{future directions}.

For this survey, we reviewed a total of 74 medical VLP papers from 2018 onwards, after collecting and filtering from Google Scholar, PubMed, and IEEE Xplore. Preprints not yet published in conferences or journals were also reviewed to cover the latest developments in the field. We considered works that utilized 2D and 3D images, as well as video, within the vision modality, and various medical reports, captions, and tweets for language. The most recent search was conducted in October 2023.

\section{Medical Multimodal Learning}
\label{multimodal learning}

Medical data can be captured, perceived, or conveyed in various forms of representation or modalities, such as image, text, video, audio, and more. Multimodal learning involves the utilization of multiple modalities of data to learn coherent knowledge. Each modality presents unique characteristics and features that complement each other.
For example, in Alzheimer's diagnosis, both FDG-PET scans and structural brain MRI provide complementary information, with FDG-PET scans revealing hypermetabolism cues and structural MRI scans detecting hippocampal atrophy \cite{chetelat2018multimodal}.

Following~\cite{zong2023self}, we formally define a unimodal labeled dataset as $D_u = \{(X_i^1, y_i^1)_{i=1}^n\}$, where $X$ is the input, $y$ is the ground-truth label, and $n$ refers to the number of samples. A multimodal dataset that contains $k$ modalities can be denoted as $D_m = \{(X_i^1, ..., X_i^k, y_i^1)_{i=1}^n\}$. The multimodal dataset that lacks labels $y$ can be represented as $(X_i^1, ..., X_i^k)$. In some scenarios, some input instances $X$ may not contain all $k$ modalities and occur as unpaired data.

\section {Multimodal learning approaches}
\label{multimodal learning approaches}

\subsection{Supervised Multimodal Learning}
Supervised multimodal learning requires ground truth labels, $y$ to learn a predictive model ${M_\theta}$ parameterized by $\theta$ such that:

\begin{equation}
\begin{aligned}
\theta^* = \underset{\theta}{\operatorname{argmin}} \sum_{\left(X_k^i, y_k^i\right) \in \mathcal{D}_m}
\mathcal{L}_{s l}\left(M_\theta\left(X_1^i, \ldots, X_k^i\right), y^i\right)
\end{aligned}
\end{equation}

Supervised multimodal learning approaches in the medical domain face challenges stemming from the demand for extensive labeled data. This difficulty is intensified by the necessity for domain experts to annotate medical data and the growing privacy concerns associated with making them publicly available.

\subsection{Self-supervised Multimodal Learning}

Self-supervised learning often comprises two stages: pretraining and downstream task learning. During pretraining, the self-supervised method does not rely on ground-truth labels. Instead, it employs some other form of self-generated supervision from the data as the objective function to train the predictive model $M_{\theta}$.
Mathematically, we replace the ground-truth labels $y$ in the supervised objective with pseudo-label $\hat{y}$, which is also a function of the input modalities. The model parameter is obtained by the following optimization:

\begin{equation}
\begin{aligned}
\theta^* = \underset{\theta}{\operatorname{argmin}} \sum_{\left(X_k^i, y_k^i\right) \in \mathcal{D}_m}
\mathcal{L}_{ssl}\left(M_{\theta}\left(X_1^i, \ldots, X_k^i\right), \hat{y}^i\right)
\end{aligned}
\end{equation}

After the pretraining phase, the trained model $M_{\theta}$ is utilized for learning specific downstream tasks that use ground-truth labels $y$. The pretrained model can be used as a feature extractor to directly extract feature embedding or fine-tuned to downstream tasks. As multimodal pretraining leveraging large-scale multimodal data, the pretrained model captures vast general knowledge that can be easily leveraged to learn task-specific models. Additionally, the self-supervised pretraining step reduces reliance on labeled data and has been shown to achieve superior performance compared to fully supervised models in certain scenarios.

In the medical domain, where obtaining labeled data is a significant obstacle, self-supervised learning proves to be highly effective. Despite the lack of annotated labels, extensive medical datasets often contain both images and detailed textual reports, offering valuable resources for self-supervised pretraining. Our focus lies in self-supervised pretraining for medical vision-language tasks, utilizing medical images and textual reports. The following sections will delve into various aspects of VLP as studied in the literature.

\section{Medical VLP Objective Functions} 
\label{Medical VLP Methods}

Here, we discuss different objective functions used in the medical VLP literature. Specifically, we categorize the objectives into four major categories: masked prediction, contrastive learning, matching prediction, and hybrid objectives. We use $X_v$ to represent the vision modality and $X_t$ to represent the text modality in the upcoming subsections.
 
\begin{figure*}[h!]
\centering
\begin{subfigure}[t]{0.46\textwidth}
    \includegraphics[width=\linewidth]{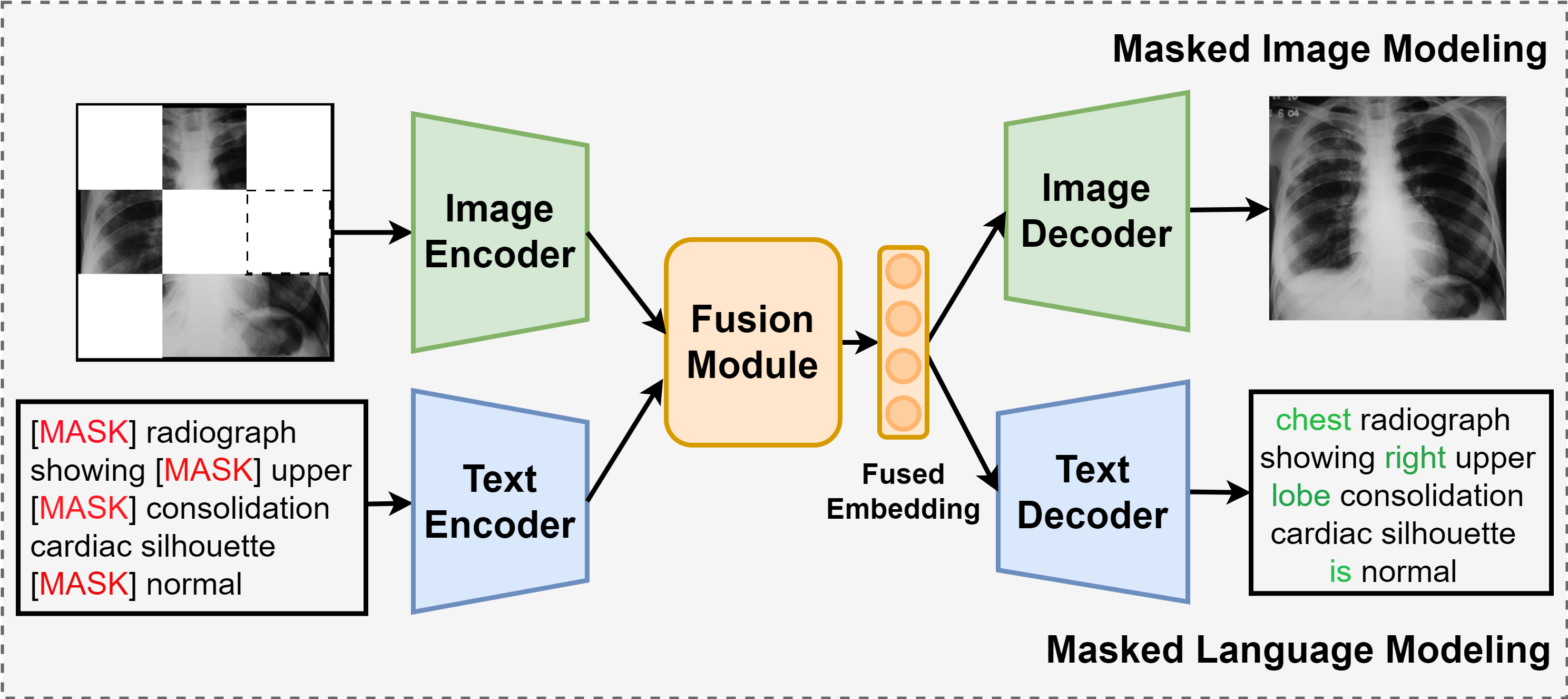}
    \caption{General Masked Prediction}
    \label{fig:masked_prediction}
\end{subfigure}
\begin{subfigure}[t]{0.33\textwidth}
    \includegraphics[width=\linewidth]{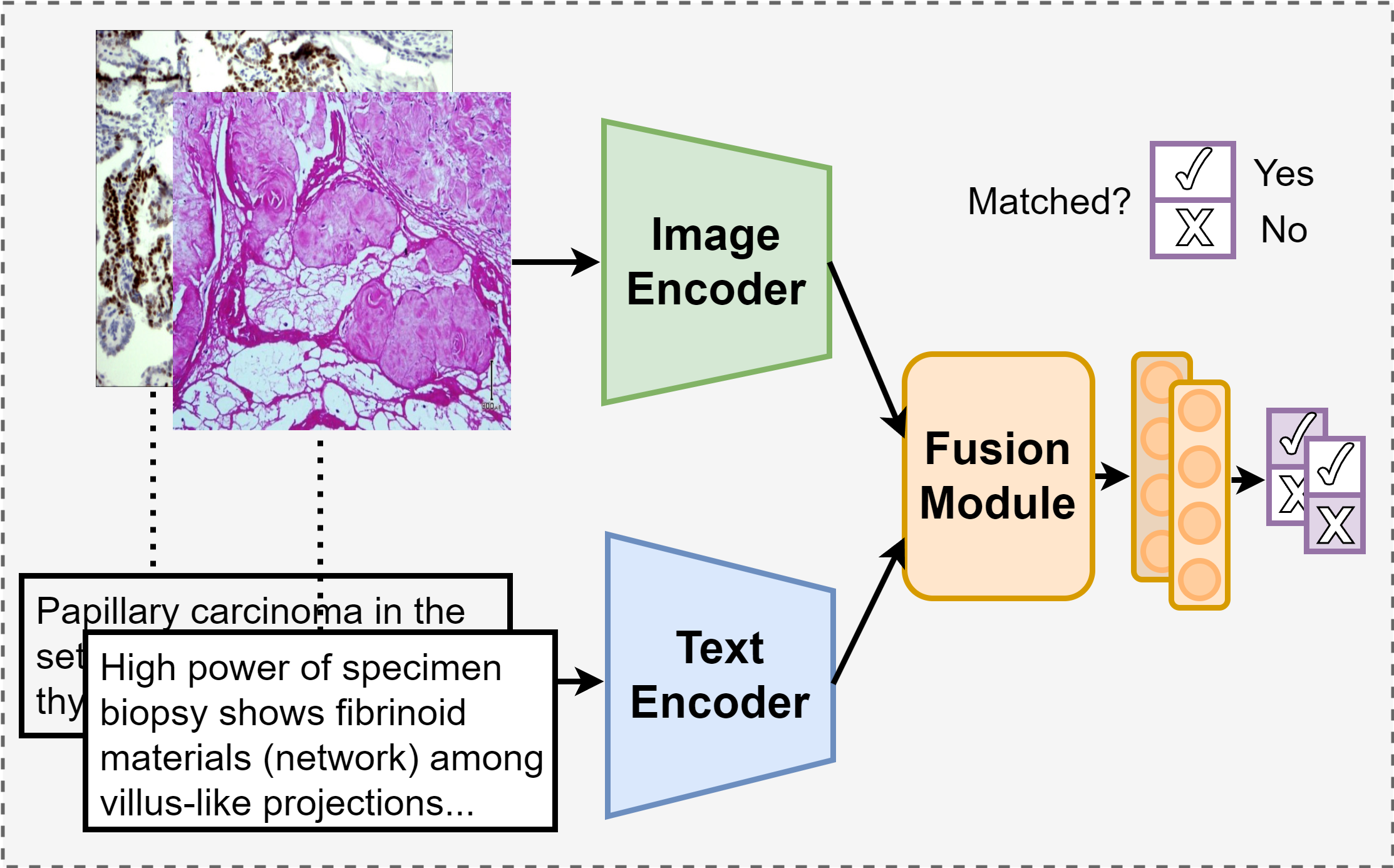}
    \caption{Matching prediction}
    \label{fig:matching_prediction}
\end{subfigure}
\hfill
\begin{subfigure}[t]{0.304\textwidth}
 \includegraphics[width=\linewidth]{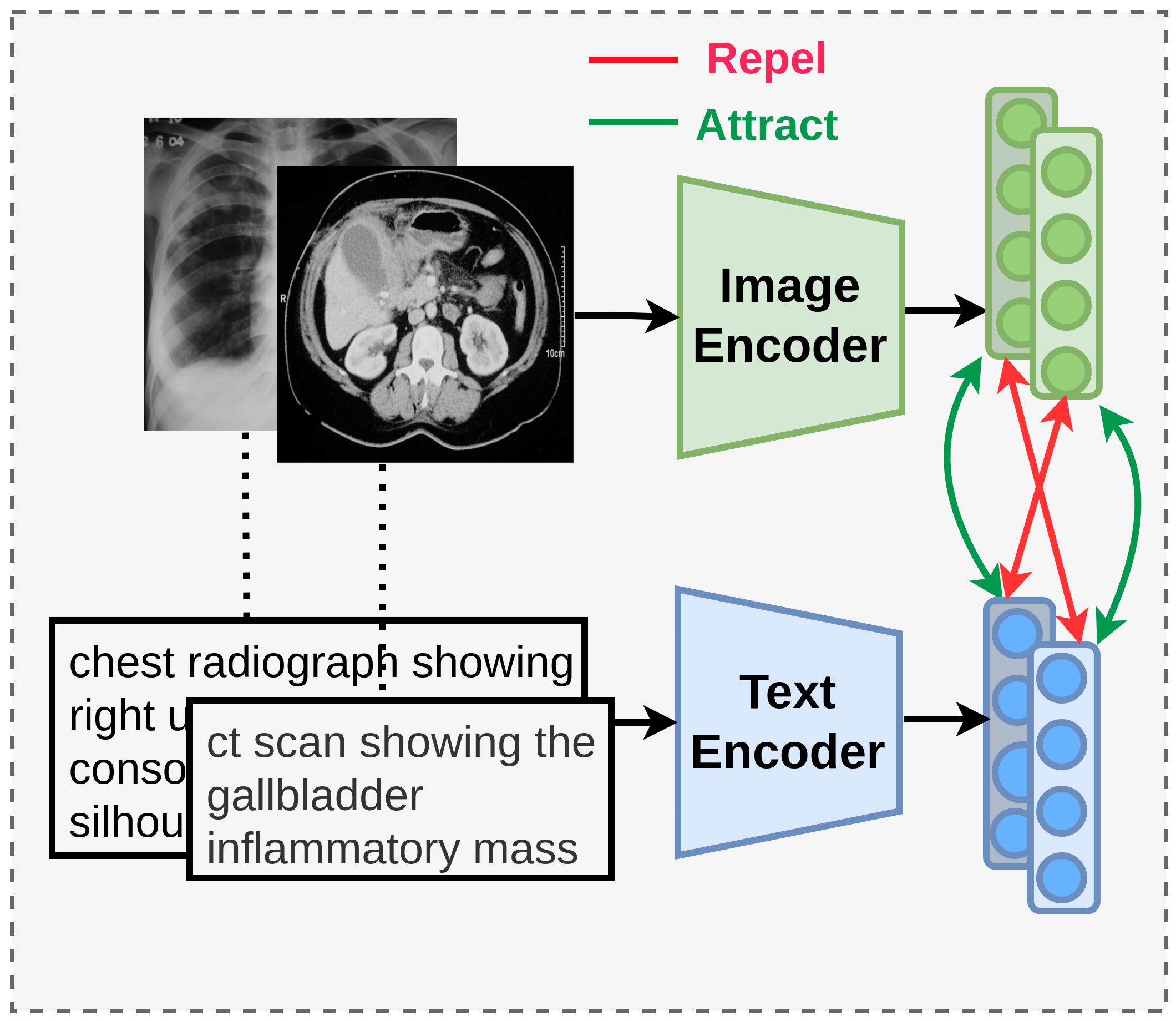}
    \caption{Global Contrastive Learning}
    \label{fig:contrastive_learning}
\end{subfigure}
\begin{subfigure}[t]{0.483\textwidth}
 \includegraphics[width=\linewidth]{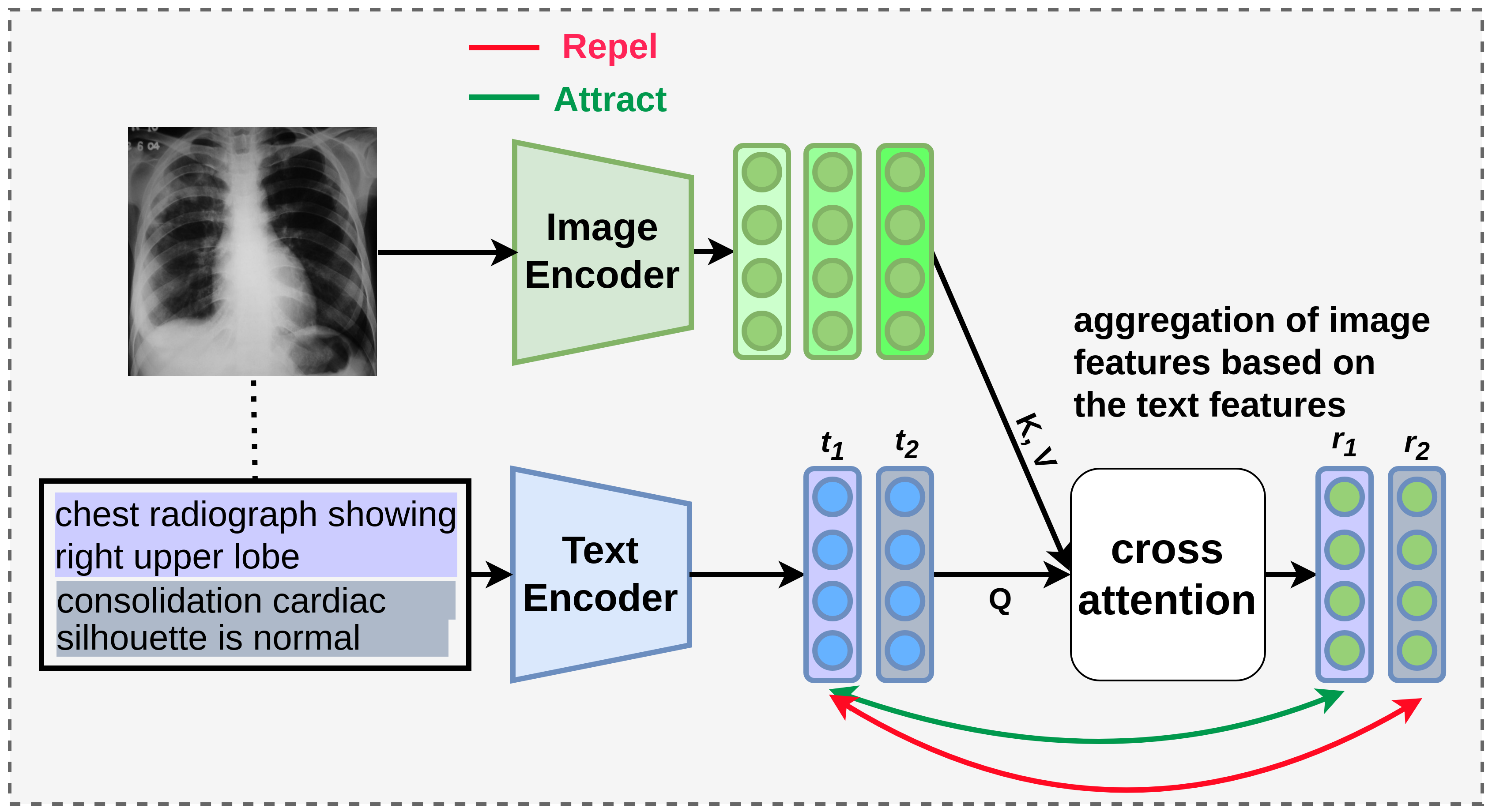}
    \caption{Local Contrastive Learning}
    \label{fig:local_contrastive}
\end{subfigure}

\caption{Pretraining Objectives} 
\label{fig:objective_function}
\end{figure*}

\subsection{Masked Prediction}

A masked prediction-based approach involves predicting or reconstructing the original, unmasked input from its masked counterpart. 
This technique employs a transformer-based encoder-decoder architecture to encode masked image/text tokens and decode the original, unmasked version. 
If $X$ represents the input and $\hat{X}$ represents the input after masking, the reconstruction loss $\mathcal{L}_{recon}$ measures the disparity between the original input and the reconstructed input of $\hat{X}$. The total loss is expressed as:

\begin{equation}
    \mathcal{L}_{mp} = \sum_{i\in(v, t)}\mathcal{L}_{recon}\left(M_\theta\left(\hat{X_v}, \hat{X_t}\right), X_i\right)
\end{equation}

As illustrated in Figure~\ref{fig:masked_prediction}, for textual data, a random subset of text tokens is masked, and the model is trained to decode the original texts; this overall process is often referred to as Masked Language Modeling~(MLM)~\cite{devlin2018bert}. Similarly, in the case of image input, a random portion of image patches is hidden, prompting the model to reconstruct the original image; while this process is called as Masked Image Modeling~(MIM)~\cite{he2022masked}. 

MMBERT~\cite{khare2021mmbert} utilized MLM for vision-language pretraining on radiograph images with corresponding captions. To achieve this, they merge image features extracted from a pretrained ResNet152 with tokenized embeddings of the accompanying text before passing them to BERT~\cite{devlin2018bert}. The model is trained with the objective of predicting masked medical text tokens using the image feature as additional context. Xu et al.~\cite{xu2023multi} also applied a similar approach by fusing image features from a pretrained Vision Transformer~(ViT) with tokenized text embeddings into a cross-modal transformer. In addition to the BERT-like objective, they employ a seq2seq objective for masked prediction~\cite{dong2019unified}, enabling its utilization in downstream medical report generation tasks. However, these methods primarily rely only on MLM objectives. MRM~\cite{zhou2023advancing} combines both MLM and MIM as pretraining objectives. Specifically, MIM is applied independently to masked low-resolution radiology images to reconstruct high-resolution versions. For MLM, the model integrates radiology image features with the text representation of heavily masked radiology report tokens to decode the input text.

Masked prediction objectives offer a straightforward approach to cross-modal interaction by conditioning one modality's reconstruction on the other. Although reconstructing raw inputs at the lowest level allows the model to learn low-level information, it introduces additional complexity due to the need for modality-specific decoders. Moreover, a slight domain gap arises between the pretraining and downstream stages, with the pretraining stage requiring a masked input while downstream tasks involve unmasked inputs. Thus, approaches relying solely on masked prediction objectives lack downstream zero-shot capabilities.

In the medical domain, datasets often encompass instances of rare diseases and conditions, typically depicted with localized and nuanced features in the images. Given that MIM relies on discerning frequent patterns for effective learning, it may encounter difficulties in accurately understanding and representing these rare occurrences. Additionally, medical texts frequently feature abbreviations and variations in writing styles, posing potential difficulties for MLM in effectively learning consistent representations.

\subsection{Contrastive Learning}
In VLP, the contrastive learning-based objective is trained with the aim of maximizing the similarity between embeddings of paired data and minimizing the similarity between embeddings of unpaired data.

\subsubsection{Cross-modal Global Contrastive Learning}
A common approach in self-supervised medical VLP methods, based on contrastive objectives, involves the use of cross-modal global alignment to maximize the similarity between global embeddings of paired visual and textual data. Using a modality-specific aggregation module denoted as $g_\phi$ and a modality-specific encoder, $E_\theta$, we obtain the global embedding $Z_i = g_{i,\phi}\left(E_{i,\theta}\left(X_i\right)\right)$, where $X_i$ represents the input data. For a given anchor $X^a$ from one modality, the paired sample $X^{+b}$ from the other modality is considered the positive sample, while all other samples are considered negative samples $X^{-b}$. The corresponding global representations of the anchor, positive sample, and negative sample are denoted as $Z^a$, $Z^{+b}$, and $Z^{-b}$, respectively. The cross-modal $a$ to $b$ global contrastive loss $\mathcal{L}_{con, a\to b}$ is commonly calculated as:

\begin{equation}
\small
    \mathcal{L}_{con, a\to b} = -\sum_i\log\frac{e^{sim\left(Z_i^a, Z_i^{+b}\right)}}{e^{sim\left(Z_i^a, Z_i^{+b}\right)} + \sum_{j=1, j\neq i}^{m} e^{sim\left(Z_i^a, Z_j^{-b}\right)}}
\end{equation}
where $sim$ represents a similarity function such as cosine similarity. The cross-modal contrastive objective can be applied in two different directions, depending on which modality the anchor sample is selected from. This objective is also commonly referred to as Image-Text Contrastive Learning (ITC) or global alignment in the literature. We illustrate this process in Figure~\ref{fig:contrastive_learning}.

ConVIRT~\cite{zhang2022contrastive} introduced a global contrastive learning method using bidirectional InfoNCE loss between image and text features for medical VLP. It utilized average-pooled embeddings from the output of the ResNet50 backbone as the global image feature and the text embedding from the BERT encoder as the text feature. CLIP~\cite{radford2021learning}, a popular model originally introduced for VLP with natural image-text pairs, shares a similar objective with ConVIRT but calculates global features using attention pooling.

Several subsequent works have effectively leveraged bidirectional cross-modal contrastive loss on medical data~\cite{ding2023improving, eslami2023pubmedclip, huang2023visual, dadoun2023joint, pan2022vision, serieys2022text}. PubMedCLIP~\cite{eslami2023pubmedclip} investigated the effectiveness of CLIP-based pretraining on medical data for medical VQA while
BiomedCLIP~\cite{zhang2023large} explored different encoder and hyperparameter choices to best adapt CLIP to medical domain. Huang et al.~\cite{huang2023visual} finetuned the general domain CLIP on a pathology image-text dataset and achieved improved results on downstream pathology datasets whereas Dadoun et al.~\cite{dadoun2023joint} applied ConVIRT on abdominal ultrasound images.

Contrastive learning paradigms suffer from feature suppression wherein the competing features cause insufficient modeling of the different potentially useful task-specific features~\cite{chen2021intriguing, pan2022vision}. A few works address this issue in the domain of placenta analysis~\cite{pan2022vision, pan2023enhancing}. Pan et al.~\cite{pan2022vision} modified the cosine similarity function in ConVIRT to the NegLogCosh similarity function, which is more sensitive to
changes in the global representation. In a follow-up work, they introduced a feature recomposition technique on top of the ConVIRT-like objective to enhance the robustness of the feature representation, capturing and representing all placental features equally~\cite{pan2023enhancing}.

Works with global alignment have mostly ignored the inherent hierarchical structure of clinical reports. IMITATE~\cite{liu2023imitate} argued that directly aligning the image with the entire report can cause misalignment, as the 'Findings' section of the report predominantly describes the image content, and the 'Impressions' section reflects and concludes the report. A Clinical Informed Contrastive Loss(CICL) was proposed that aligns the high-level visual features and impression section, and the multi-level visual features and findings section.

While the global contrastive learning loss offers a straightforward framework for VLP, it is often suboptimal for localized downstream tasks such as semantic segmentation and object detection. Given the localized nature of abnormalities in medical image-text pairs, these learned global representations are insufficient to properly represent the localized visual cues~\cite{huang2021gloria}. Objectives that additionally enforce local alignment between word-image regions or sentence-image regions and learn local representations explicitly have demonstrated improved performance on these localized tasks and also provide improved interpretability~\cite{muller2022joint, huang2021gloria}. We discuss local alignment in the next section.

\subsubsection{Global and Local Contrastive Learning}
Considering the sequence of text and visual vectors generated from a paired sample, denoted as $\{t_1,\ldots ,t_m\}$ and $\{v_1,\ldots ,v_n\}$ respectively, we compute a new set of text-based aggregated visual vectors as $\{r_1,\ldots, r_m\}$ where $t_i = f\left(t_i,  \{v_1,\ldots ,v_n\}\right)$, 
The general text-to-vision local alignment loss is then calculated as follows:
\begin{equation}
\small
    \mathcal{L}_{local, t\to v} = 
    -\sum_N\sum_{i=1}^{m}\log\frac{e^{sim\left(t_i, r_i\right)}}{e^{sim\left(t_i, r_i\right)} + \sum_{j=1, j\neq i}^{m} e^{sim\left(t_i, r_j)\right)}}
    \label{eqn:local_alignment}
\end{equation}
Simply put, in text-to-vision local alignment, the text-fragment~(word or sentence) feature and the aggregated visual features based on the same text-fragment are brought closer, whereas those corresponding to different text fragments are pulled apart~(Figure~\ref{fig:local_contrastive}).
In general, local alignment is performed within the features of a paired image-text sample, whereas global alignment occurs between features across different pairs.

GLORIA~\cite{huang2021gloria}, unlike ConVIRT, introduced a local contrastive loss in addition to the global alignment objective. For this, each word embedding is aligned with attention-weighted image features using a bidirectional contrastive loss. This resulted in an improved performance across multiple tasks including segmentation, while providing the learned attention weights for model interpretability. Assuming that medical images often present consistent visual structure, LIMITR~\cite{dawidowicz2023limitr} additionally added 2D positional encodings to the image features and utilized both frontal and lateral Chest X-ray views to improve pretraining. Meanwhile, LRCLR~\cite{rizvi2023local} argued that medical information often manifests in small regions within the image. For that, it utilizes the self-attention weights of the image encoder to select these local regions and employs a cross-modal transformer to contextualize image regions into text embedding. Finally, a local contrastive loss is then computed between the class token embeddings of the image and text.

Works like GLORIA, LRCLR, and LIMITR perform local alignment at the level of word and image region features. 
However, words alone lack proper context and are not guaranteed to have direct visual correspondence, possibly resulting in misalignments~\cite{liu2023imitate}. Assuming a sentence in a clinical report to be better associated with localized image regions, Liao et al.~\cite{liao2021multimodal} proposes to maximize the Mutual Information~(MI) between the sentence and the image region pair with the highest MI Estimation. LOVT~\cite{muller2022joint} additionally encourages nearby image region representations to be similar by allowing multiple positives per region, benefiting localized downstream tasks.

The addition of local alignment objective provides image region representations that better capture the localized pathological visual cues present in medical datasets. Such learned representations are beneficial for localized downstream tasks such as object detection and segmentation~\cite{muller2022role}. Additionally, the learned attention weights used in the aggregated feature calculation provide a means of visualizing the connection between image regions and words/sentences which is not possible solely with global contrastive loss. 

Local alignment with medical data however does pose some unique challenges. Works on local representation learning in the natural domain often utilize pretrained object detectors or ground truth annotations for supervision, which is difficult to obtain in the medical domain. Thus the local representations and alignments have to be learned by relying only on instance-level pairing information which is a much weaker form of supervision. Additionally, contrastive objectives in the natural domain have been shown to benefit when the number of negatives is increased~\cite{chen2020simple}. Since the contrastive alignment objective in equation~\ref{eqn:local_alignment} adopted by most works in this category only utilizes the features of the same paired image-text sample, the number of negatives available is limited. The limitation can be mitigated by increasing the number of local features per sample or using local representations across samples. However, the effect of negatives on local alignment and downstream performance in medical VLP requires further investigation~\cite{muller2022joint}.

\subsubsection{False negatives in global contrastive learning}

Contrastive learning is prone to false negatives, wherein the negative samples may belong to the same category as the anchor or share similar semantics even if they come from different samples. This issue is more problematic in the medical domain, where the dataset is often highly imbalanced, 
with a very small number of rare pathologies compared to common diseases or healthy samples.
As a result, there is a higher likelihood that negative samples may belong to the same category as the anchor. Additionally, medical datasets are often multi-class in nature, with each case involving multiple pathological conditions~\cite{jang2022significantly}, and thus the anchor and the negative samples may share some abnormality conditions. These false negatives inadvertently push the representations of data with similar semantics apart, producing sub-optimal image-text representations~\cite {jang2022significantly}.

To mitigate the impact of false negatives in global alignment, Jang et al.~\cite{jang2022significantly} proposed an approach to relax the similarity function of the contrastive loss of CLIP by clipping its upper bound. This resulted in improved performance in zero-shot classification on medical data. Meanwhile, ReCO~\cite{lin2023relaxing} does not penalize orthogonal or negatively correlated negative pairs to keep the embedding space flexible. Some works proposed the use of domain knowledge to address the false negatives issue. For example, MedCLIP~\cite{wang2022medclip} used UMLS~\cite{bodenreider2004unified} to build a knowledge-driven similarity label that softens the semantic matching loss. This can also be viewed as a regularization of the contrastive loss based on semantics. KoBo~\cite{chen2023knowledge} also proposed the use of domain knowledge to mitigate the problems of false negatives and the inconsistency between semantics and text morphology due to the use of biased expressions and negative expressions by radiologists.

Contrastive objectives are simple and computationally inexpensive due to uncomplicated architecture without complex decoders and fusion modules. However, the noise and bias introduced by false negatives remain an open problem that needs further study in the medical context.

\subsection{Matching Prediction}

Matching prediction is trained with the objective of predicting whether the input vision and text pairs match~(positive pairs) or not~(negative pairs)~(Figure~\ref{fig:matching_prediction}). For a given vision-text pair, the model should output a high probability if both the input vision and text modality belong to the same paired instance and a low probability if not. The objective is formulated as a binary cross-entropy loss:
\begin{equation}
\begin{aligned}
    \small
    \mathcal{L}~(X_v, X_t) = \:&\mathbb{E} [\hat{y} \cdot \log(M_\theta(X_v, X_t)) + \\
    &(1 - \hat{y}) \cdot \log(M_\theta(X_v, X_t))]
\end{aligned}
\end{equation}
where $\hat{y}$ is a pseudo label with a value of 1 if the vision and text modality belong to the same pair and 0 if not. In medical VLP, this matching prediction objective is commonly referred to as Image-Text Matching~(ITM) or Image-Report Matching~(IRM). Typically, matching prediction is used in conjunction with other pretraining objectives~\cite{moon2022multi, chen2022multi, chen2022align}, as discussed in Section~\ref{Hybrid with matching}.

\subsection{Hybrid}
\label{Hybrid}
While individual pre-training techniques exhibit strong performance, they may still exhibit certain limitations. For instance, contrastive and matching prediction methods are prone to false negatives, while masked modeling methods lack zero-shot capabilities. 
Many approaches choose to combine multiple pretraining objectives to achieve complementary effects. A composite objective encompassing M individual objectives can be expressed as a weighted sum:

\begin{equation}
    \mathcal{L} = \sum_{i = 1}^{M} \lambda_{i} \mathcal{L}_i
\end{equation}

\subsubsection{Masking + Contrastive}
The learning principles of contrastive and masking-based strategies complement each other. Cross-modal contrastive learning explicitly discriminates positive and negative vision-text pairs, enhancing the discriminative capabilities of the representations. On the other hand, masked prediction objectives promote a more fine-grained representation learning and can better capture low-level modality information~\cite{chen2023contrastive, cheng2023prior}. We also note approaches utilizing unimodal variations of these objectives in this section.

 A few works have been proposed combining cross-modal contrastive learning and masked prediction on joint features during pretraining. To improve the modeling of cross-modal interactions, PMC-CLIP~\cite{lin2023pmc} included the MLM objective on the fused image-text embedding in addition to the global alignment used by CLIP. Meanwhile, MPMA~\cite{zhang2023multi} applied global and local alignment objectives in addition to the masking-based objectives of MRM~\cite{zhou2023advancing}, and Silva et al.\cite{silva2023contrastive} introduced a supervised contrastive los--mitigating the effect of false negatives--in addition to the masking-based objectives of MMBERT. PRIOR\cite{cheng2023prior} also proposed using both local and global alignment, alongside cross-modality conditional reconstruction objectives of masked images and reports. Additionally, it utilizes prototype vectors to represent sentence features, guided by the insight that clinical descriptions can be effectively replaced by structured labels alone, eliminating the need for syntax information.

Some works also utilize unimodal contrastive learning (where alignment is performed between augmentations of the same modality input) or unimodal masked modeling on unpaired medical data, to initialize the encoders before pretraining. For instance, Liu et al.~\cite{liu2023medical} and BioViL~\cite{boecking2022making} utilized unimodal MLM to train the text encoder and unimodal contrastive learning to train the image encoder before utilizing cross-modal contrastive losses during VLP. Additionally, considering the inherent structure of clinical reports, they aligned the representation of different sections of the report. 

The localized visual cues in medical images tend to be masked when applying masked prediction on the visual modality. A masked medical image can be ambiguous and no longer perfectly aligned with the accompanying text. Applying contrastive learning naively to masked inputs can thus provide suboptimal benefits due to the misalignment. Most works bypass this issue by applying masking and contrastive objectives separately, either in different iterations/phases ensuring that alignment is performed on representations of unmasked inputs. A computationally simpler approach can be applying strategies to mitigate the misalignment while applying the two objectives together. Such strategies include adjusting the contrastive loss based on the location of masked image patches~\cite{huang2023enhancing} or reconstructing the features of masked patches before alignment~\cite{chen2023contrastive}. However, the performance gains with such strategies have been marginal, and utilizing the location information alone without the semantic content is not suitable when dataset variability is high, necessitating more robust strategies.

\subsubsection{Masking + Matching prediction}
\label{Hybrid with matching}
Since contrastive learning is performed with little to no modality fusion, it is unable to model rich interactions between modalities, whereas, matching prediction is applied on fused multimodal embeddings and is better equipped to capture the cross-modal interactions. 

Many approaches have incorporated a combination of masked prediction and matching prediction objectives for pretraining. UWOX~\cite{wang2021self} proposed a versatile method capable of pretraining on paired, unpaired~(image only or text only), and a combination of paired and unpaired image-text datasets. The method utilizes modality-specific encoders with shared weights, a cross-correlation module to align image and text features, and modality-specific decoders. Additionally, pretraining involves MIM, Multi-Scale Masked Vision Modeling, and matching prediction losses. Meanwhile, PTUnifier~\cite{chen2023towards} proposed a unified architecture that uses prompts to handle the paired and unpaired inputs. It uses MLM, ITM, and ITC objectives for training. 

MedViLL~\cite{moon2022multi} used MLM and Image Report Matching~(IRM) objectives for pretraining. 
Additionally, they proposed a new self-attention mask, Bidirectional Auto-Regressive~(BAR), that combines the strengths of both bidirectional~\cite{devlin2018bert} and seq2seq masking schemes ~\cite{vaswani2017attention}.
M3AE~\cite{chen2022multi} used MLM, ITM, and MIM objectives employing transformer-based modules while ARL~\cite{chen2022align} proposed injecting medical domain knowledge into VLP by utilizing the UMLS knowledge base. To achieve this, they employed knowledge-guided image and text similarity alignment, multi-modal fusion by knowledge, and knowledge-aware ITM and MIM objectives.  

\subsubsection{Mixed Objectives}
    
    A few works utilize a combination of masking, contrastive and matching prediction losses for pretraining. Li et al.~\cite{li2023self} proposed to use MIM, MLM, ITM, and Image text alignment via contrastive learning for pretraining on medical image caption datasets and evaluated on medical VQA.
    MUMC~\cite{li2023masked} applied both unimodal and multimodal contrastive losses, in addition to MLM and ITM objectives, and demonstrated improved performance on VQA. It utilized MoCo-based momentum encoders for unimodal contrast, and random image patches were masked as an augmentation approach.

Beyond these objectives, various other objectives—such as generation, clustering, distillation, and their combinations—are explored in the literature to enhance pretraining. Addressing the challenge of limited large-scale pretraining datasets in the medical domain,~\cite{seibold2022breaking} proposed using a distillation-based SiamSiam loss to align image representations between augmentations. Simultaneously, they employed global and local contrastive losses to align image and text features. For the local contrastive loss, they utilized a modification of MIL-NCE~\cite{miech2020end}, a multiple-instance learning-based objective. This approach is based on the insight that not all sentences in the report describe the image. However, it can be assumed that all clinically relevant information about the image is contained within a subset of the sentences.

 To reduce false negatives in contrastive learning, a clustering objective can be introduced that takes semantic similarity into account by grouping semantically similar samples into the same cluster. For instance, to leverage disease-level semantic correspondence across examples, MGCA~\cite{wang2022multi} performed alignment at three different levels: the global instance level, the local pathological region level, and, additionally, at the disease~(prototype) level. For the disease-level alignment, the soft cluster assignment of one modality, provided by the Sinkhorn-Knopp clustering algorithm, is used to train the representation of the other modality.

REFERS~\cite{zhou2022generalized} utilized a report generation objective on the image embeddings in addition to image-to-text contrastive loss. Additionally, it proposed incorporating multiple views of the patient when available to improve image representation. Meanwhile, TIER~\cite{palepu2023tier} introduced a regularization scheme. Based on the observation that text tokens should describe only a small number of image regions, and each image region should correspond to only a few text tokens, TIER employs a regularization that penalizes the entropy of text tokens concerning image patch similarity scores.
UniCLAM~\cite{zhan2022uniclam} introduced adversarial masks designed to maximize the unimodal contrastive loss. By trying to minimize the discriminability between the masked and unmasked input, the adversarial masks tend to mask semantic regions in the input and improve interpretability.

Using a combination of methods can be beneficial, however, special care should be taken while adjusting hyperparameters for training as different pre-training objectives introduce additional complexity to the training process and may introduce objective interference, where optimizing one objective undermines another.

\subsubsection{Hybrid with additional classification objective}
Some approaches have been proposed that additionally introduce a classification objective in the pretraining, injunction with other mixed objectives. These approaches utilize additional labels present in the dataset or labels extracted using external tools to supervise the objective. 
Primarily, these approaches aim to make use of additional supervision obtained from external labeling tools and knowledge bases to enhance pretraining.

For instance, Clinical-BERT~\cite{yan2022clinical} proposed a multi-label classification task on the CheXpert labels present in the MIMIC-CXR~\cite{johnson2019mimic} dataset in conjunction with different domain knowledge enhanced masked prediction and alignment objectives.
Meanwhile, UniBrain~\cite{lei2023unibrain} included a multi-class classification objective using a transformer decoder architecture that takes in the aligned MRI image features as key and value and a set of brain disease descriptions as the queries to predict existence labels for each disease. This approach enables zero-shot disease classification using the disease descriptions derived from UMLS~\cite{bodenreider2004unified}.

MedKLIP~\cite{wu2023medklip} proposed a triplet extraction module to exclusively extract medical-specific information from clinical reports and then encode the triplets using domain knowledge bases. It utilizes contrastive and classification objectives for training where classification is performed by treating the triplet values as labels. 
KAD~\cite{zhang2023knowledge} proposed the use of domain knowledge and available off-the-shelf tools to enhance medical VLP and facilitate zero-shot image classification. Specifically, KAD first trains a knowledge encoder using contrastive learning by sampling positives and negatives from the UMLS system 
which is then used to encode the processed report.
In addition to using contrastive loss for aligning the image and report embeddings, for each report, a set of existence labels against the top $Q$ entities in the whole corpus is used as labels for the classification objective.

Objectives utilizing classification objective necessitates labels extracted from the clinical reports. Extracting labels involves not just identifying pathology entities amidst their diverse expressions but also accurately recognizing and quantifying negation and uncertainty. However, the tools used for label extraction and negation detection may be restricted in their capabilities and coverage. For example, the CheXpert labeling tool~\cite{irvin2019chexpert} and the Radgraph tool~\cite{jain2021radgraph} for entity extraction and determining their presence are specific to chest radiograph reports and are not well suited for reports from modalities like CT and MRI or different anatomical locations. Thus the applicability of this objective can depend upon the available tools for that domain.

\section{Medical VLP Related Additional Aspects }
In this section, we have discussed a few particularities in the medical VLP domain, not covered in previous sections.

\subsection{Using unpaired data for pretraining}
    The effectiveness of self-supervised learning is directly correlated with the amount of pretraining data available. 
    To overcome the limited size of medical paired image text reports, many papers have proposed methods that utilize unpaired datasets during pretraining. We refer to image-only or text-only datasets, with or without other accompanying annotations, as unpaired datasets.
    
    A number of works utilize the class labels present in labeled image datasets to create a text sentence using prompts. For instance,
    CXR-CLIP\cite{you2023cxr} and Seibold at al.~\cite{seibold2022breaking} proposed using static prompts to generate additional paired data from a labeled image dataset.
    Meanwhile, UCML\cite{wang2023unified} suggested using both learnable continuous and static discrete prompts to bridge the gap between the label and the medical report inputs.
  Silva et al.~\cite{silva2023foundation} generated text reports from category labels using expert knowledge to form image-text pairs from 37 fundus imaging datasets and applied them to create a pretrained model.

    Meanwhile, other works have employed unpaired data without labels using generative approaches. 
   For example, GTGM~\cite{chen2023generative} proposed generating reports from images using a report generator module fine-tuned on MediCaT while ~\cite{liu2023utilizing} suggested using clinical reports only and generated synthetic images. They used RoentGen~\cite{chambon2022roentgen}, a pretrained diffusion model adapted to generate synthetic Chest X-rays conditioned on MIMIC-CXR radiology reports. The synthetic dataset, including the radiology reports, was used for pretraining a model that achieved comparable results to methods trained on the real dataset. Additionally, MEDIMP~\cite{milecki2023medimp} utilized clinical tabular data to generate vocabulary-rich text reports from template sentences and LLM.

\subsection{Using temporal information during pretraining}
    Unlike typical natural language text, clinical reports frequently incorporate references to past history, providing a temporal context. For instance, the phrase~\textit{"Chest X-ray showing a decrease in consolidation and broncho-vascular markings following treatment"} encompasses both the associated pre-treatment and the post-treatment image. Additionally, such descriptions can be ambiguous and can refer to any instance with consolidation when the temporal connection is ignored~\cite{bannur2023learning}. BioViL-T~\cite{bannur2023learning} proposed using prior images and reports, when available, to harness the temporal supervision present in medical datasets for VLP.

\subsection{Use of multiple views of Chest X-ray}
\label{different_views}
In clinical practice, clinicians often rely on insights from multiple views of radiology scans for accurate assessment. Distinct visual cues tend to be more apparent on different views, and leveraging these diverse perspectives, when available, can enhance the learned representation and prove beneficial in downstream evaluation tasks.
LIMITR~\cite{dawidowicz2023limitr} proposed using the lateral view of the chest X-ray alongside the frontal view when available. Since LIMITR uses both global and local alignment, a combined feature from both views is employed for the global alignment, while preserving individual region features for local alignment. Similarly, REFERS~\cite{zhou2022generalized} also unified the multiple Chest X-ray views into a single representation, improving the downstream performance.

\subsection{Other visual modalities}
    Previous sections primarily focused on works involving 2D images as the predominant visual modality. Here, we delve into specific works related to other visual modalities, such as 3D images and surgery videos.

Relatively fewer works have been proposed in the case of 3D medical VLP.
    MedBLIP~\cite{chen2023medblip} and Niu et al.~\cite{niu2023ct} adapted a 2D ViT image encoder to process 3D inputs by slicing 3D volume into smaller volumes and flattening each smaller volume into a long input vector. 
    Other works, such as MEDIMP~\cite{milecki2023medimp} and UniBrain~\cite{lei2023unibrain}, directly used 3D ResNet encoders to process the input 3D volume. Meanwhile, Van et al.~\cite{van2023exploring} performed domain-adaptive pretraining to 2D patches sampled from 3D lung CT scans which are either obtained using zero-shot cross-modal retrieval using CLIP or random sampling.

    SurgVLP~\cite{yuan2023learning} extended medical VLP to the video domain. They segmented surgical lectures into random short clips and extracted audio transcripts for each clip using ASR models. Their pretraining approach involved aligning the video clips with corresponding transcribed sentences.

\section{Data Augmentation}
Augmentation is an integral part of medical VLP as it helps improve the robustness, generalization, and performance of models for diverse data. As the number of data in the medical domain is particularly limited, augmentation can greatly help to increase the size and variability of the dataset. In this section, we briefly mention commonly used augmentation techniques in medical VLP.

\subsection{Image Augmentation}
Image augmentations involve the use of different transformations such as resizing of the image, random cropping, rotation, random affine transformation, and flipping, or manipulations such as color jitter, random grayscale, and random masking. 
Proper consideration should be taken while applying spatial augmentation, such as flipping, cropping, and masking, as there can be misalignment between the augmented image and the spatial description in the report. For example, the report may describe the condition in the left lung, and after horizontal flipping, the image may no longer be aligned with the description. Similarly, cropping and masking can remove important pathological cues from the image as described in the accompanying text.

\subsection{Text Augmentation}
It has been shown that applying text augmentation can be beneficial towards the performance on vision language tasks~\cite{fan2023improving}. 
The text augmentation techniques that are primarily used include shuffling of sentences~\cite{you2023cxr, boecking2022making}, random sentence sampling~\cite{zhang2022contrastive}, and back translation~\cite{you2023cxr}.
Shuffling of sentences involves randomly shuffling sentences of findings and impression section of the report and are based on the assumption that reports of radiological findings are usually permutation invariant~\cite{boecking2022making}. Meanwhile, in sentence sampling, a sentence is randomly selected from a multi-sentence report and treated as a whole text. Back translation utilizes machine translation models to translate the original sentence to a target language and back, introducing variations in the original text. Windsor et al.~\cite{windsor2023vision} also utilized nearest-neighbor search as in the De-CLIP framework~\cite{li2021supervision}, to find similar reports as text augmentation.

\section{Architecture}
\label{architecture}
\subsection{Encoder Architecture}
Here, we briefly discuss various architectures used for the image and text encoders in medical VLP.

\subsubsection{Image Encoder}
\begin{figure}[h!]
\centering
\begin{subfigure}[t]{0.24\textwidth}
    \includegraphics[width=\linewidth]{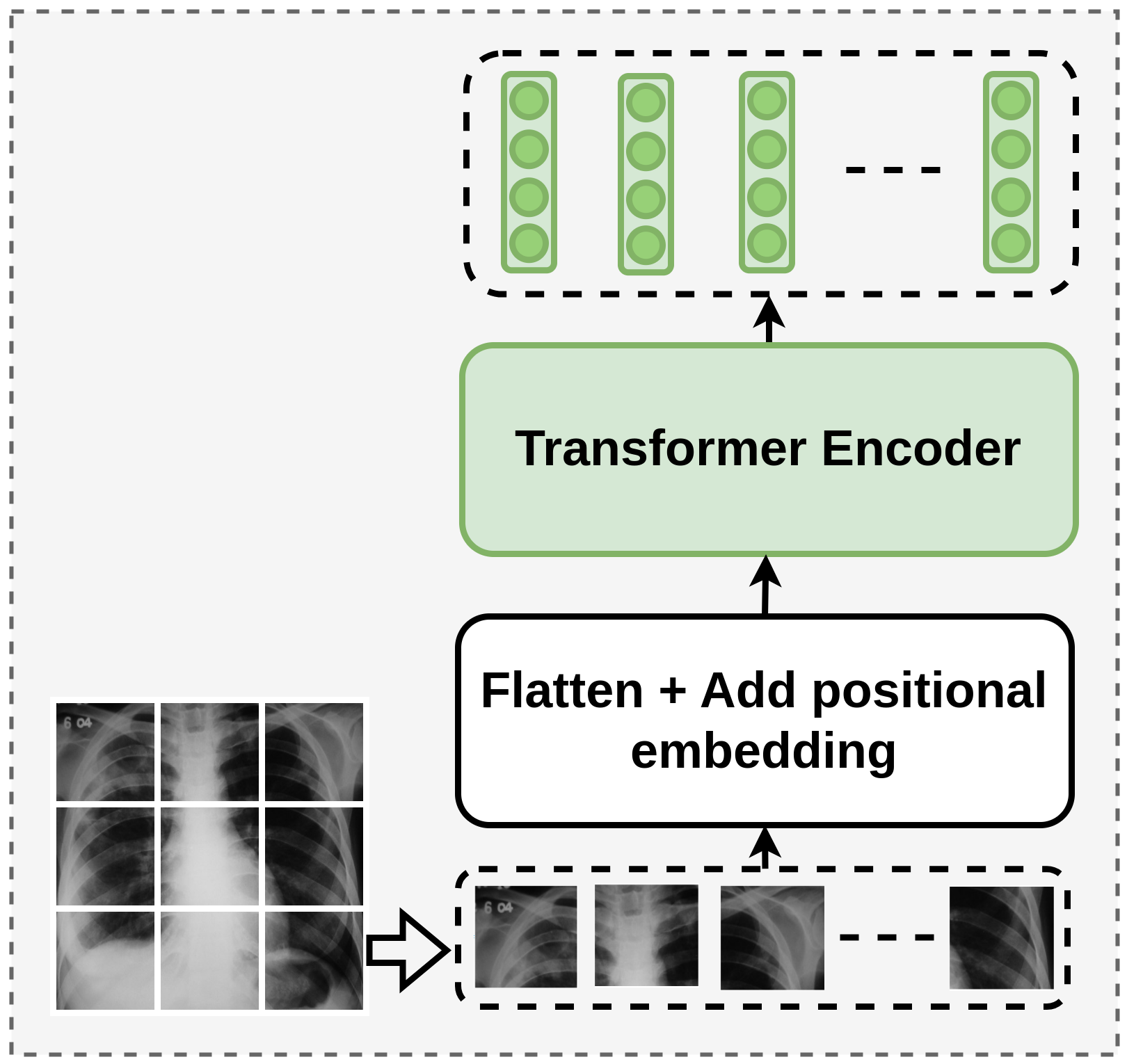}
    \caption{ViT}
    \label{fig:vit}
\end{subfigure}
\begin{subfigure}[t]{0.24\textwidth}
    \includegraphics[width=\linewidth]{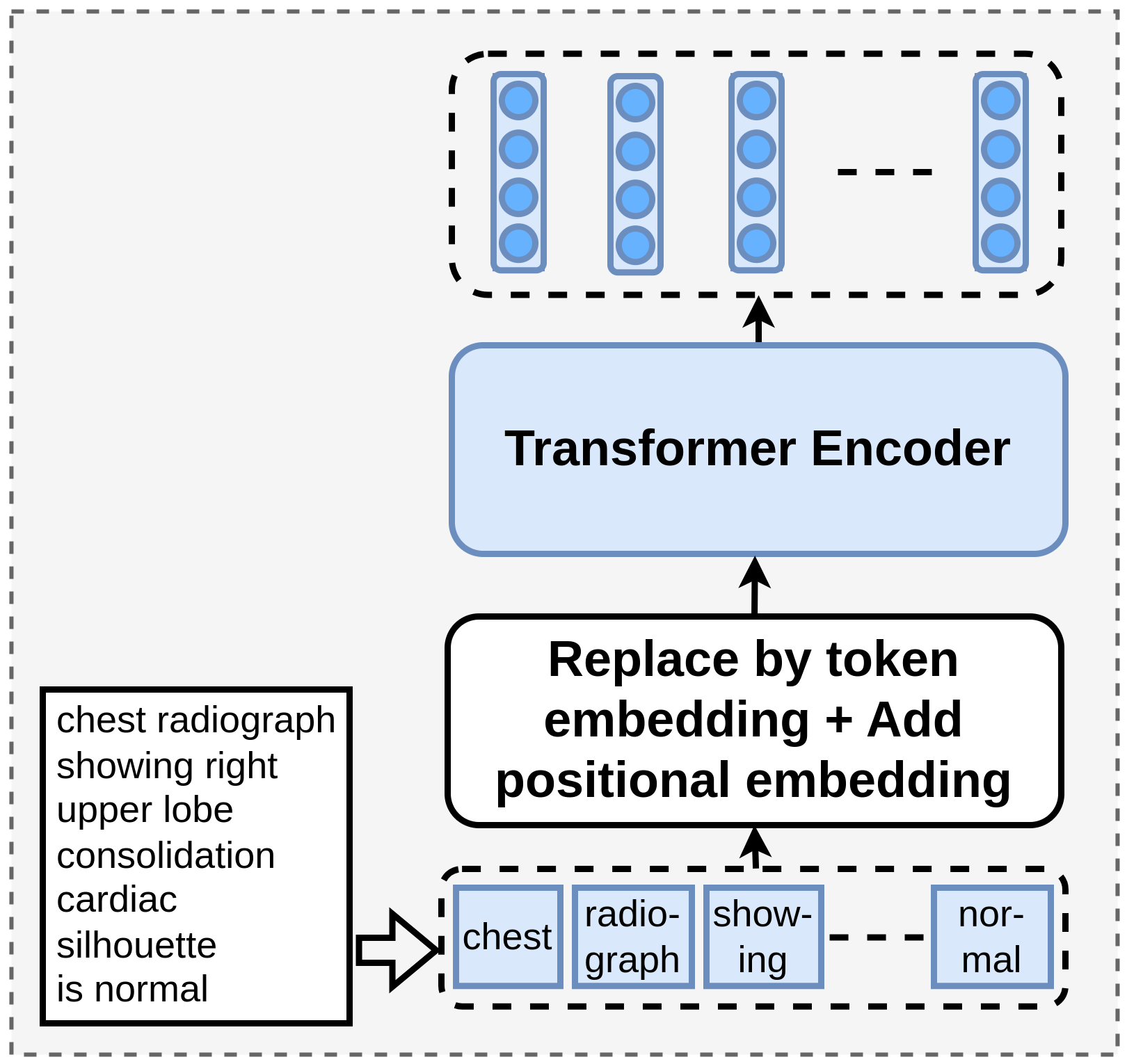}
    \caption{BERT}
    \label{fig:bert}
\end{subfigure}
\begin{subfigure}[t]{0.483\textwidth}
    \includegraphics[width=\linewidth]{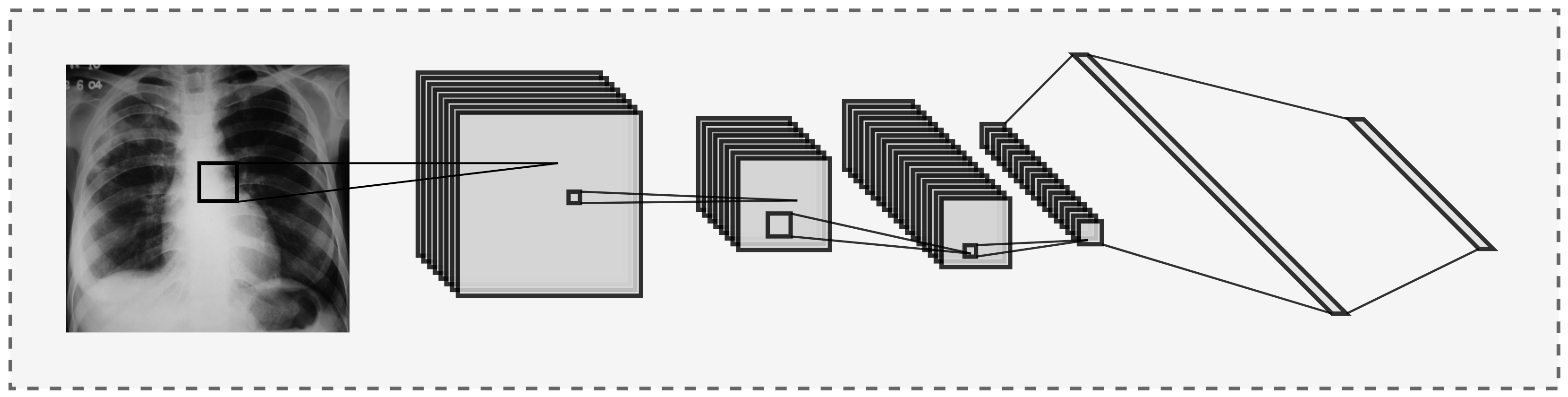}
    \caption{CNN}
    \label{fig:cnn}
\end{subfigure}
\caption{Encoder Architectures} 
\label{fig:encoders}
\end{figure}

The image encoder employs different CNN and ViT architectures. As depicted in Figure~\ref{fig:encoders}, a CNN utilizes layers of convolution and pooling operations to process the visual input~\cite{lecun1998gradient}, while a ViT breaks down the input image into a number of non-overlapping image patches and employs several layers of self-attention layers to contextualize the patch-level features~\cite{dosovitskiy2020image}. ResNet50~\cite{he2016deep} is the most commonly used CNN architecture, while a few works have also employed other ResNet encoders and the EfficientNet model~\cite{silva2023contrastive}. Similarly, ViT-B/16 and ViT-B/32~\cite{dosovitskiy2020image} are the most commonly used ViT architectures. The Swin Transformer~\cite{liu2021swin} has also been found to be used by some approaches~\cite{wang2022medclip, you2023cxr}. Approaches trained on MIM, where masking is performed at the patch level, utilize ViT architectures. Regarding encoder initialization, some approaches initialize the encoder with pretrained weights in the natural domain, such as the image encoder of CLIP~\cite{radford2021learning} or ResNet trained on ImageNet. A few approaches also propose pretraining the image encoder on medical data using domain-specific augmentations, which has been observed to benefit downstream tasks~\cite{boecking2022making, shu2023miter, windsor2023vision}.

\subsubsection{Text Encoder}
All the works covered in this survey have utilized transformer-based architectures for the text encoder. Similar to ViT, a transformer-based text encoder breaks down the input text into a number of smaller units called text tokens. Multiple layers of self-attention are used to process and contextualize the token-level features~\cite{devlin2018bert}. Most approaches employ a BERT-based architecture as the text encoder, with weights initialized from publicly available BERT weights pretrained on clinical text~\cite{gu2021domain, alsentzer2019publicly, windsor2023vision}. Encoders initialized in this way are better equipped to handle the text semantics of clinical reports. Notably, certain works, such as BioViL~\cite{boecking2022making}, go a step further by pretraining text encoders with in-domain text datasets themselves, enhancing their adaptability to the specific domain.

\subsection{Modality Fusion}

\begin{figure}[h!]
\centering
\begin{subfigure}[t]{0.23\textwidth}
    \includegraphics[width=\linewidth]{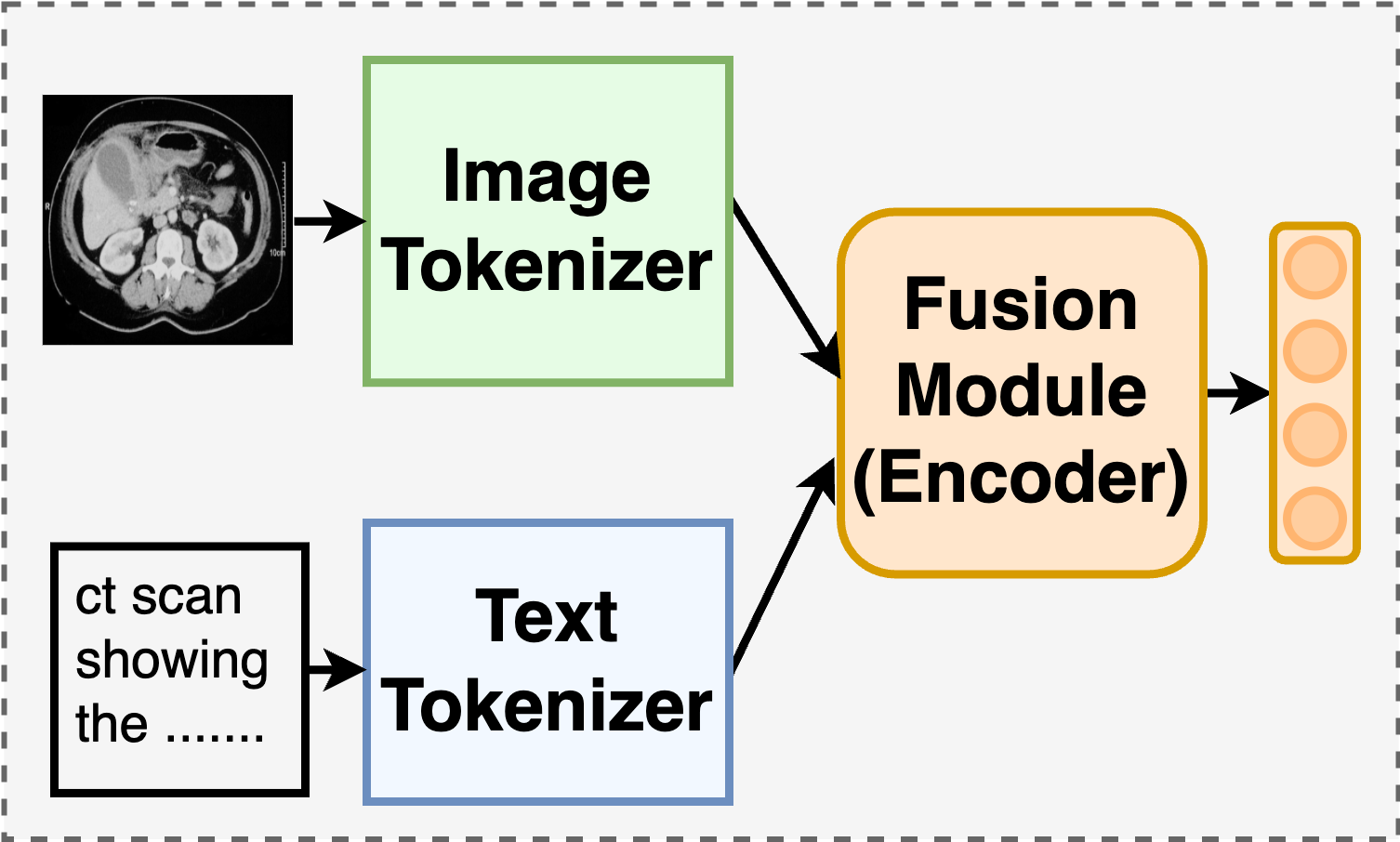}
    \caption{Early Fusion}
    \label{fig:early_fusion}
\end{subfigure}
\hfill
\begin{subfigure}[t]{0.25\textwidth}
    \includegraphics[width=\linewidth]{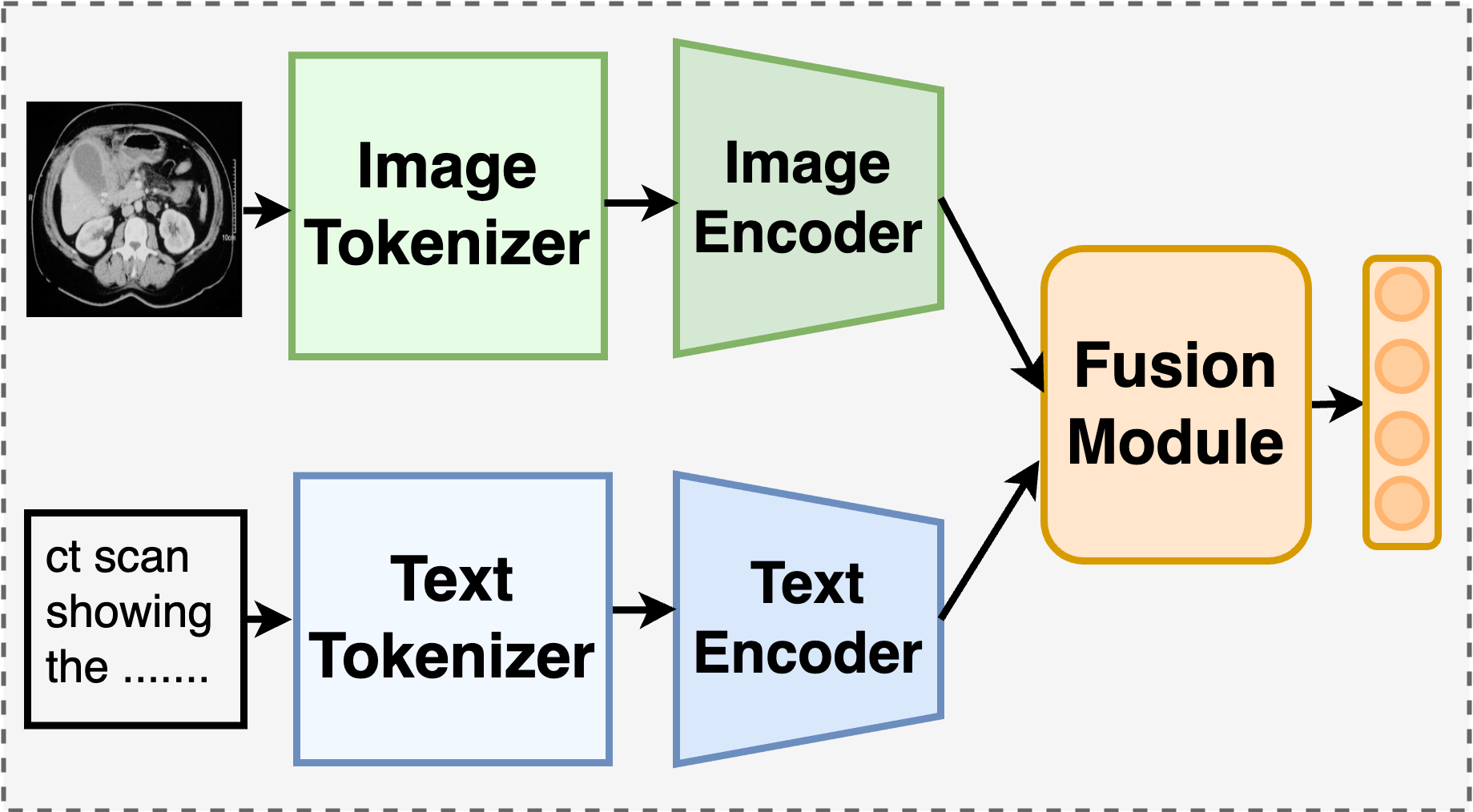}
    \caption{Late Fusion}
    \label{fig:late_fusion}
\end{subfigure}
\caption{Fusion Approaches} 
\label{fig:fusion}
\end{figure}

Modality fusion involves combining and integrating information from multiple modalities. In VLP, a predictive model $M_\theta$ comprises various components, including modality-specific tokenizers, modality-specific encoders, a fusion module that merges modality-specific encoded features, and an objective-specific prediction head. The fusion module plays a pivotal role in consolidating different modality features into a shared embedding. In this section, we explicitly distinguish the tokenizer from the encoder for enhanced clarity. While the architectures of both may be similar, they serve distinct purposes. A tokenizer dissects the raw input into low-level tokens/features for input to transformer-based encoders, whereas an encoder (whether CNN or transformer-based) utilizes tokens/raw inputs to derive a meaningful numerical representation. In this section, we consider various engineering decisions involved in combining visual and text modality features during pretraining and broadly categorize them into modality-specific encoders without fusion, early fusion, and late fusion.

\subsubsection{Modality-specific encoders without fusion}
These approaches do not fuse the different modalities and instead only align them using contrastive objectives~\cite{zhang2022contrastive, wang2022medclip, zhou2022generalized, zhang2023knowledge, wang2023unified, eslami2023pubmedclip, huang2023visual, zhang2023large}. All approaches that only perform global alignment fall under this category. Due to the lack of fusion modules and subsequent decoder/prediction heads, such architectures tend to be computationally simpler. 
Additionally, the lack of fusion implies that such approaches can be easily adapted for downstream unimodal and cross-modal tasks but tend to be limited for multimodal tasks such as VQA~\cite{chen2023towards}, which requires both text and image as inputs.

\subsubsection{Early fusion}
As illustrated in Figure~\ref{fig:early_fusion}, we categorize methods that combine the two modalities using a common encoder module into this category. Such architectures can also be referred to as single-stream architectures with unified encoders. Methods that utilize CNNs or ViTs for image tokenization are also included here if the text modality does not require a separate encoder before fusion. For instance, MMBERT~\cite{khare2021mmbert}, Clinical-BERT~\cite{yan2022clinical}, and MedViLL~\cite{moon2022multi} utilize CNN modules to generate image tokens, which are then passed alongside the text tokens through a unified transformer encoder. Meanwhile, Xu et al.~\cite{xu2023multi} used the Swin Transformer for image tokenization. The several layers of self-attention provide contextualized multimodal embeddings when trained with self-supervised objectives.

\subsubsection{Late fusion}
As shown in Figure~\ref{fig:late_fusion}, methods that fuse the different modality information after the use of separate encoders belong to this category. Such approaches can also be referred to as dual-stream architectures.

For late fusion, attention-based fusion modules are popularly used, perhaps due to their effectiveness in modeling long feature sequences and the minimal engineering requirements for effective multimodal interaction. For instance, M3AE~\cite{chen2022multi}, ARL~\cite{chen2022align}, and PTUnifier~\cite{chen2023towards} used several layers of cross-attention to fuse the image and text embeddings, while PMC-CLIP~\cite{lin2023pmc} applied several self-attention layers to the concatenation of image and text embeddings to employ the MLM objective. Meanwhile, MUMC~\cite{li2023masked} and~\cite{li2023self} used alternating self and cross-attention layers to fuse the image and text embeddings. MPMA~\cite{zhang2023multi}, in its cross-attention layers, additionally utilized learnable matrices to represent memory elements, aiming to enhance cross-modal fusion.

Some works have also employed alternative forms of fusion. MITER~\cite{shu2023miter} used linear sum to fuse aligned image and text features for the ITM prediction head, while in PRIOR~\cite{cheng2023prior}, text-weighted image features and masked image features are concatenated for the reconstruction objective. Meanwhile, MRM~\cite{zhou2023advancing} and CMITM~\cite{chen2023contrastive} fused the two modalities by adding global average pooled image features to each of the text token embeddings before feeding the fused embeddings to the text encoder. We categorize MRM and CMITM as late fusion, as the CNN used for obtaining image embeddings acts as an encoder and not a tokenizer.

Other works applying the classification objective primarily used the transformer decoder architecture as the fusion module, with the image features as the input key and value embeddings. For instance, MedKLIP~\cite{wu2023medklip} and KAD~\cite{zhang2023knowledge} use entity embeddings as the query, whereas UniBrain~\cite{lei2023unibrain} uses disease description embeddings.
We also categorize approaches trained with local alignment and utilizing the other modality features to calculate the aggregation weights~\cite{huang2021gloria, boecking2022making, muller2022joint, wang2022multi, dawidowicz2023limitr} into late modality fusion. 

\section{Downstream Evaluation Tasks}
\begin{figure*}[h!]
\centering
\begin{subfigure}[t]{0.41\textwidth}
    \includegraphics[width=\linewidth]{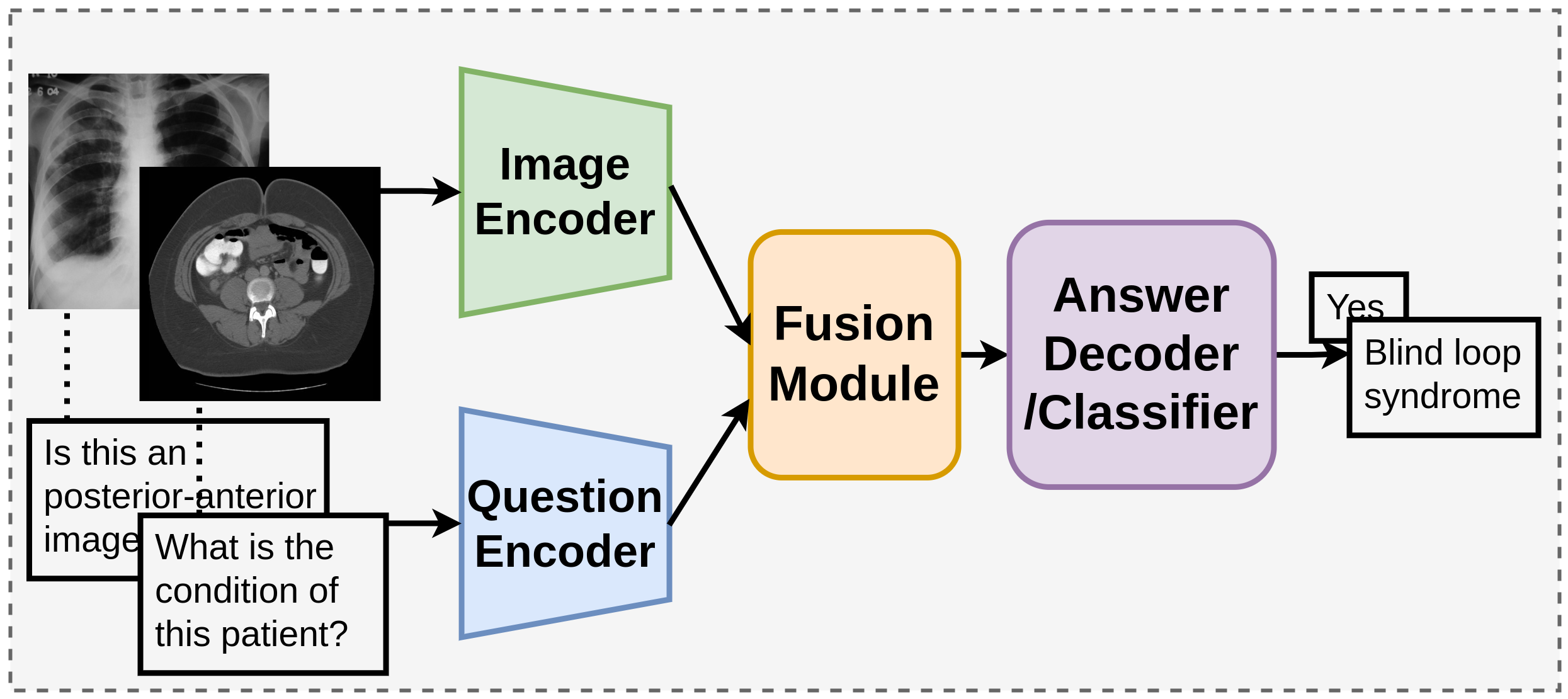}
    \caption{Visual Question Answering~(VQA)}
    \label{fig:vqa}
\end{subfigure}
\begin{subfigure}[t]{0.36\textwidth}
    \includegraphics[width=\linewidth]{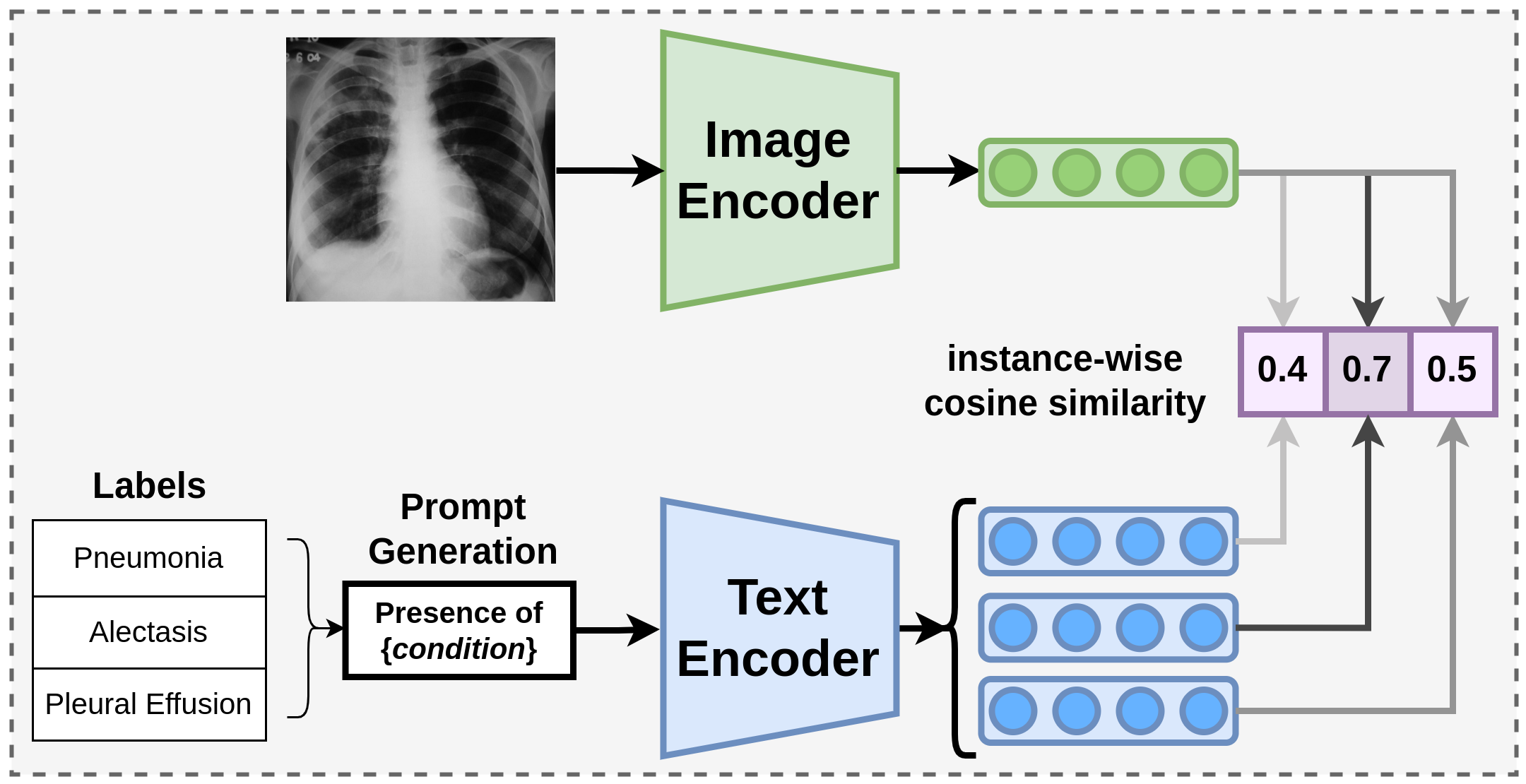}
    \caption{Zero-shot Classification}
    \label{fig:zero_shot_classification}
\end{subfigure}
\begin{subfigure}[t]{0.35\textwidth}
    \includegraphics[width=\linewidth]{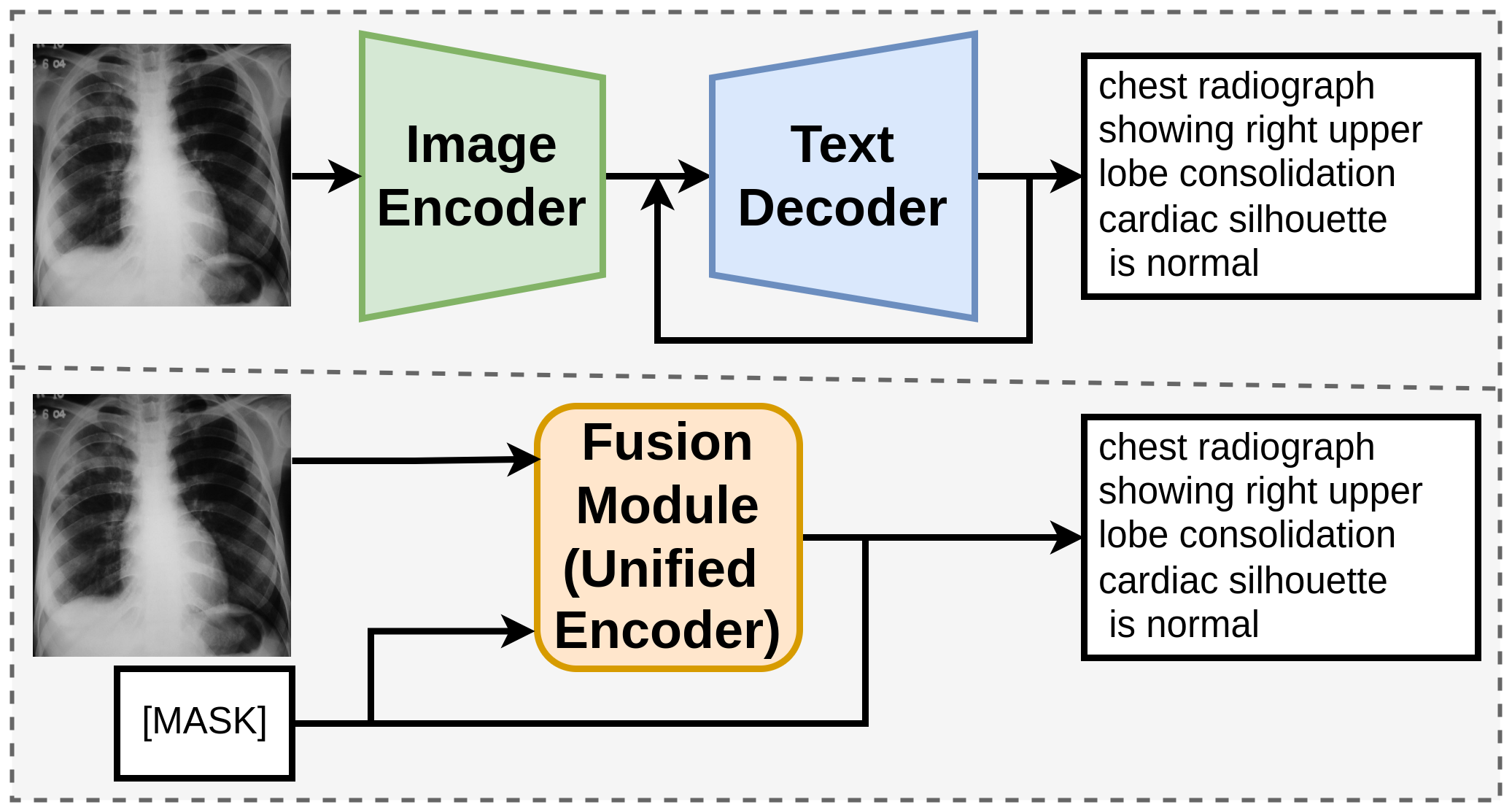}
    \caption{Report Generation}
    \label{fig:report_generation}
\end{subfigure}
\begin{subfigure}[t]{0.42\textwidth}
    \includegraphics[width=\linewidth]{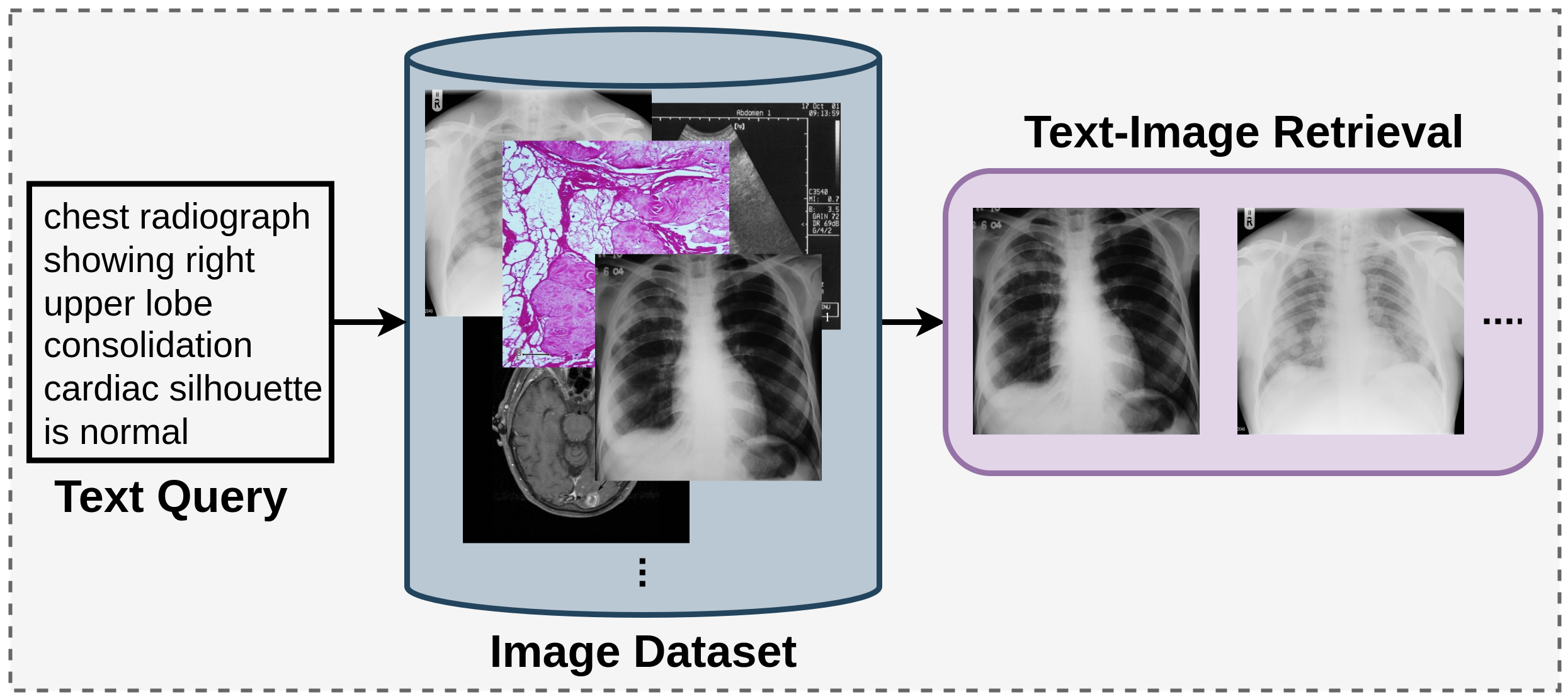}
    \caption{Cross-modal Retrieval}
    \label{fig:retrieval}
\end{subfigure}
\caption{Illustration of some Downstream Evaluation Tasks} 
\label{fig:downstream_tasks}
\end{figure*}
\label{downstream evaluation tasks}
A wide array of medical downstream tasks can be improved through vision-language pretraining. The pretrained model incorporates a wealth of medical knowledge from both vision and language sources, enhancing its feature representation and transferability. The downstream tasks can directly leverage this generalized knowledge either by initializing the downstream model with pretrained weights or by extracting features directly from the pretrained model. In this section, we explore various medical downstream tasks that can be enhanced using vision-language pretraining and how different works have been evaluated and adapted to these tasks.

\subsection{Medical Image Classification}
Medical image classification involves the task of assigning either a single label or multi-labels to a medical image. Evaluation in classification typically relies on test accuracy, while additional metrics such as F1-score, Sensitivity, Specificity, etc., are often considered to provide nuanced insights into the classification performance. For binary and multi-label classification, metrics like AUC-ROC, Matthews Correlation Coefficient~(MCC), Precision, and Recall are commonly employed. Pre-trained models can be leveraged for classification tasks in various ways: 1) zero-shot classification through prompting; 2) linear probing on extracted features; and 3) fine-tuning the entire pre-trained model for classification.

    \subsubsection{ Zero shot classification~(prompting)}
    
    Zero-shot classification in VLP is a fairly recent concept that can directly utilize the pretrained model to make inferences on downstream datasets without requiring additional training. To achieve this, a list of medical labels, or prompts, representing specific pathological conditions is generated from the medical text vocabulary. Then, for a given image, an image encoding is compared against the text encoding of the list of prompts, both generated using a CLIP-like VLP architecture. The text encoding corresponding to a prompt with the highest cosine similarity with the given image provides the test label for that input (Figure~\ref{fig:zero_shot_classification}).

    GLORIA~\cite{huang2021gloria} and CheXZero~\cite{tiu2022expert} are early works that evaluated their models pretrained on the MIMIC-CXR dataset to perform zero-shot image classification using static prompts in the medical domain. Following this, several works have employed VLP in zero-shot medical image classification~\cite{huang2021gloria,wang2022medclip,seibold2022breaking,boecking2022making,wu2023medklip,chen2023knowledge,wang2023unified,jang2022significantly,silva2023foundation, dack2023empirical}. However, static prompts like those in CheXZero may fail when the target label is absent in the pretraining dataset. Qin et al.~\cite{qin2022medical} investigated the effectiveness of transferring VLM pretrained in the general domain to the medical domain using prompts and found that employing expressive prompts describing visual characteristics is crucial for better generalization of the pretrained models. Additionally, they proposed an automated prompt generation strategy using pretrained language models and/or VQA models on the medical VL dataset. CLIPath~\cite{lai2023clipath} investigated zero-shot classification capabilities of CLIP in the medical domain, proposing a parameter-efficient finetuning approach.

    Zero-shot classification is highly effective in scenarios where labeled data is limited in the downstream task. However, VLP trained with masked prediction objectives alone lacks zero-shot classification capabilities and cannot be directly adopted for downstream tasks. Therefore, an additional pretraining objective involving contrastive learning is often needed.

    \subsubsection{Linear Probing}

    Linear probing involves training only a linear classification head on the downstream labeled dataset, utilizing features extracted from the pretrained model, with the encoder layers held frozen. This approach serves to evaluate the generalizability of the pretrained model's knowledge without extensive retraining. Several VLP approaches have tested their pretrained models on downstream medical classification tasks using linear probing~\cite{wang2021self,huang2021gloria,wang2022medclip,zhang2022contrastive,wang2022multi}. Linear probing is typically employed in scenarios where the downstream dataset domain aligns with the pretraining dataset. However, in cases where a significant domain gap exists between the downstream and pretraining datasets, this approach may not be the best. For instance, a model trained on MIMIC-CXR~\cite{johnson2019mimic}~(a chest X-ray pretraining dataset) will perform poorly on MELINDA~\cite{wu2021melinda}~(containing figures of biomedical experiments), but can perform well on CheXpert~\cite{irvin2019chexpert}~(a chest X-ray classification dataset).

    \subsubsection{Fine tuning}
    
    This evaluation process enables the entire model to adapt its learned representations to better align with the nuances and specific patterns present in the new dataset or task. In this context, both the encoder and a linear classification head undergo training on the downstream datasets, a strategy well-suited for addressing domain gaps between the pretraining and downstream datasets. Several pretrained VL models have undergone assessment using this approach~\cite{huang2021gloria,wang2022multi, lei2023unibrain, boecking2022making,wang2022medclip} 

    An alternative approach for transformer-based encoders involves fine-tuning for downstream classification tasks without changing the pretrained encoder. In this case, the pretrained encoder remains frozen, and tunable prompt tokens are introduced to facilitate adaptation without altering the stability of the pretrained encoder~\cite{jia2022visual}. Such an approach enables parameter-efficient fine-tuning. Zhang et al.~\cite{zhang2023text} applied Visual Prompt Tuning~(VPT)~\cite{jia2022visual} to fine-tune a pathological image dataset.

\subsection{Medical Image Segmentation} 
Medical image segmentation aims at identifying region within an image that relates to distinct organs/tissues/cells, diseased area, or some other region of interest. This task requires a model to capture local context and structures at the pixel level, making it inherently challenging. The performance of a model on medical image segmentation is generally evaluated using measures such as DICE scores, Contrast-to-Noise Ratio~(CNR), and mean Intersection over Union~(mIoU), against the ground truth segmentation mask~(often manually annotated regions). A pretrained VL model that has contextualized local information through medical texts is better suited to boost the downstream segmentation task.
\subsubsection{Zero-shot segmentation}
In zero-shot segmentation, the pretrained model can be directly employed to perform region segmentation without fine-tuning the pretrained model. However, adapting to a downstream task with zero-shot segmentation is often challenging, given that the objective function for training the segmentation task differs from that of VLP models. Nevertheless, BioViL~\cite{boecking2022making} proposed using the attention weights learned during local alignment to conduct zero-shot semantic segmentation on RSNA Pneumonia. Similar to zero-shot classification approaches, a text prompt is created based on the segmentation target category. Then, a similarity score between the prompt's representation and all the grid-level image features is computed to determine the binary label for each grid, providing a coarse-grained prediction for segmentation. The coarse-grained segmentation prediction map is compared against the bounding box level annotation in the dataset to evaluate performance. However, the use of grid-level predictions may be too coarse-grained, and obtaining pixel-level predictions with this strategy could be challenging.
\subsubsection{Fine-tuned segmentation}
Most works that evaluate image segmentation involve fine-tuning under different data settings~\cite{zhou2023advancing, chen2023knowledge,poudel2023exploring}.  A common approach for fine-tuning pretrained models on segmentation includes using pretrained image encoder weights to initialize the frozen encoder of U-net~\cite{ronneberger2015u} and fine-tuning only the decoder section\cite{huang2021gloria, wang2022multi, wu2023medklip, liu2023m, liu2023imitate,muller2022radiological}, while some fine-tune the encoder as well~\cite{muller2022radiological}.

\subsection{Medical Object Detection}
Object detection involves identifying and locating objects within a given image. The location of these objects needs to be accurately outlined using bounding boxes. Object detection is primarily evaluated using mean Average Precision~(mAP). 

A general object detection fine-tuning framework consists of an encoder and a detection head. During fine-tuning, the encoder is initialized with the weights of the pretrained image encoder. Most of the methods use YOLOv3 detection head~\cite{wang2022multi},~\cite{muller2022joint},~\cite{liu2023m},~\cite{muller2022radiological},~\cite{liu2023utilizing}, while PRIOR~\cite{cheng2023prior} used R-CNN detection head.
Some of the methods fine-tune only the detection head~\cite{wang2022multi, muller2022joint, liu2023m},~\cite{muller2022radiological, cheng2023prior, liu2023utilizing}, while some fine-tune both backbone and detection head~\cite{muller2022joint,muller2022radiological}.

We also consider Phrase Grounding, where a text phrase is associated with an image region, to be a multimodal variation of classical object detection. Phrase grounding is generally used to evaluate the performance of local alignment in a zero-shot manner. 
Similar to the zero-shot segmentation approach, zero-shot grounding has been performed by utilizing the grid-level attention weights between the text phrase and image regions~\cite{dawidowicz2023limitr, pan2023enhancing, boecking2022making}.
To evaluate the performance of the different prompt fusion techniques, Guo et al.~\cite{guo2023multiple} performed zero-shot and few-shot grounding for lesions using GLIP~\cite{li2022grounded} framework, using the dot product of region features and phrase features for prediction. 

\subsection{Cross-modal Retrieval}

Cross-modal retrieval involves the retrieval of information across different modalities, such as retrieving relevant data from one modality based on a query from a different modality (Figure ~\ref{fig:retrieval}). Depending upon the dataset and formulation of retrieval a single query may have one or more relevant counterparts. Retrieval is typically evaluated using Recall@$n$ and Precision@$n$ measures which measure the accuracy of the model in obtaining the correct result out of top $n$ retrievals. 
Pretrained models have been adapted for retrieval in both zero-shot and fine-tuned manner in the literature.

\subsubsection{Zero shot retrieval}
Contrastively trained models can be easily adapted to zero-shot retrieval by relying on the similarity between the query and possible candidates. Similar to zero-shot classification using prompts, the candidate whose embedding results in the highest cosine similarity with the query embedding is taken as the best candidate. Multiple predictions are ranked based on this similarity value when $n > 1$
~\cite{huang2021gloria,zhang2022contrastive,wang2022medclip,lin2023pmc,chen2023knowledge,yuan2023learning,shu2023miter,cheng2023prior,wang2023using, you2023cxr}.

Works trained with the ITM loss are also capable of performing zero-shot retrieval. For instance, matching scores obtained from the ITM head can be used to rank the candidate queries and performed retrieval in a zero-shot manner~\cite{chen2022multi, chen2022align, chen2023towards}

Although models trained using ITM also perform well on zero-shot retrieval tasks, their computational demands soar when dealing with a large number of candidates. This arises from the necessity to calculate the fused embedding for each query-candidate pair, which results in NxM forward passes, where N is the number of queries and M is the number of candidates. In contrast, contrastive objectives require only N+M forward passes to compute all necessary embeddings, making them more computationally efficient for retrieval with large datasets.

\subsubsection{Fine-tuned retrieval}
For fine-tuning, MedViLL~\cite{moon2022multi}, M3AE~\cite{chen2022multi}, ARL~\cite{chen2022align} and PTUnifier~\cite{chen2023towards} initializes the similarity score head with weights of the pretrained ITM head and then trains with binary cross-entropy loss to classify matched and mismatched pairs. Meanwhile, UWOX~\cite{wang2021self} applied a fully connected layer to combine the image and text embeddings and trains by Cauchy hashing loss.

While contrastive and ITM objectives allow for strong zero-shot performance, fine-tuning is beneficial when the evaluation dataset differs from the pretraining dataset~\cite{moon2022multi}.

\subsection{Medical Report Generation}
Report generation involves the automatic generation of descriptive reports given a medical image as input. Most works are evaluated on report generation using Natural Language Generation(NLG) metrics such as BLEU, METEOR,
CIDEr, and ROUGE-L. Some works~\cite{endo2021retrieval, yan2022clinical} argue that NLG metrics might not be sufficient in medical report generation and that prioritizing the accuracy of diagnoses within the reports could be more crucial than solely focusing on generating reports with similar vocabulary and structure. Consequently, these methodologies utilize external labeling tools like CheXbert~\cite{smit2004chexbert} to compare diagnosis labels between reference and generated reports and use clinical efficacy metrics such as Precision, Recall and F1 scores in addition to the NLG metrics. However, the effectiveness of such clinical efficacy metrics could be limited by the coverage of these labeling tools and might not be universally applicable across diverse medical datasets.

\subsubsection{Zero shot generation}
We found a single work that adapted pretrained models for zero-shot report generation without requiring fine-tuning with autoregressive objectives.
CXR-RePair~\cite{endo2021retrieval} proposed a retrieval-based approach to report generation where the task is reformulated as searching for a report or a set of sentences from a corpus, most similar to the given image. To address the generation time complexity due to possibly large search space, they propose to cluster the sentences and only retain representative elements in the search corpus. While the approach appears to perform poorly on the NLG metric BLUE2, results on clinical efficacy metrics that focus on the presence of correct diagnosis as present in the reference report show its effectiveness. 

\subsubsection{Fine-tuned generation}
Works such as Clinical-BERT~\cite{yan2022clinical}, MedViLL~\cite{moon2022multi}, BioVil-T~\cite{bannur2023learning} and~\cite{xu2023multi} that utilize a unified encoder with early fusion architecture, fine-tune the pretrained unified encoder with sequence-to-sequence objective for downstream report generation, thus requiring no additional modules. Each word is autoregressively generated using the previously generated words and image tokens as the context. For dual-stream architectures with separable image and text encoders fine-tuning is generally done by introducing a transformer decoder on top of the vision encoder as illustrated in Figure~\ref{fig:report_generation}~\cite{zhang2023multi, shu2023miter}.

\subsection{Medical Visual Question Answering}
Visual Question Answering revolves around analyzing an image and generating appropriate textual responses to questions posed in natural language. VQA is similar to the report generation objective in that both require the generation of text as output. However, it additionally includes the natural language question as the input alongside the image. 
Medical VQA is often formulated as a classification task where the model aims to find the correct response among a set of possible responses obtained from the training set. This is possible due to the limited possibilities present in existing VQA datasets.
Different accuracy metrics are used for the evaluation
while some methodologies also evaluate NLG metrics such as BLUE.

As shown in Figure~\ref{fig:vqa}, the general medical VQA fine-tuning framework consists of an image encoder, a text encoder to encode the question, a fusion module, and a classifier head. Most works initialize the image and text encoder with the pretrained encoders of the VLP model~\cite{lin2023pmc, shu2023miter}, while some opt for a different text encoder~\cite{zhang2023multi, eslami2023pubmedclip}. Architectures with transformer-based unified encoder instead either apply the classifier head on the class token during fine-tuning~\cite{xu2023multi} or fuse the different token features before applying the classification head~\cite{khare2021mmbert}.

\section{Datasets}
\label{datsets}

\begin{table*}[!ht]
    \centering
    \renewcommand{\arraystretch}{1.2}
    \begin{tabular}{l|l|l|l} 
        \hline
         Dataset & Image Type & Text type & Number of Samples  \\
         \hline
         MIMIC-CXR \cite{johnson2019mimic} & Chest X-ray & Medical report & 227,835 study reports with 377,110 associated images \\ 
         
         Open-I \cite{demner2016preparing} & Chest X-ray & Medical report & 3,996 radiology reports with 8121 associated images\\   
         
         Openpath \cite{huang2023visual} & Pathology images & Tweets & 208,414 image-text pairs \\ 
         
         ROCO \cite{pelka2018radiology} & Radiology scans & Caption & 81,825 image-text pairs, + 6,127 out-of-class set \\ 
         
         MedICaT \cite{subramanian2020medicat} &  Mixed modalities & Caption & 217,060 image-text pairs \\
         
         PMC-OA \cite{lin2023pmc} &  Mixed modalities  & Caption & 1,646,592 image-text pairs \\  
         
         ImageCLEF2022 \cite{de2022imageclef} & Radiology scans & Caption & 98,565 image-text pairs \\   
         
         RGC \cite{xu2023multi} & Radiology scans & Caption & 18,434 image-text pairs \\ 
         
         ARCH \cite{gamper2021multiple} & Histology & Caption & 15,164 images with 11,816 unique captions    \\ 
         
         PatchGastric \cite{tsuneki2022inference} & Histopathology & Medical report & 262,777 patches from 991 WSI image-report pairs \\ 
         
         Quilt-1M \cite{ikezogwo2023quilt} &  Histopathology & Mixed sources &  1M image-text pairs\\

         SSPH Dataset \cite{lei2023unibrain} & MRI & Report & 24,770 image-report pairs \\

         OASIS \cite{marcus2007open} & T1 MRI & Reports \footnote{Diagnosis are not made public} & 3,020 volumetric T1 MRI\\

         \hline
    \end{tabular}
    \caption{Vision-Language pretraining datasets: Image types, Text types, and Number of samples. Here, ``Mixed modalities" indicates ``Radiology, Histology, \& others", while ``Mixed sources" indicates ``Audio transcripts, Tweets, \& captions".}
    \label{pretraining_dataset}
\end{table*}

\subsection{Pretraining Datasets}

We present a list of the most commonly used image text pretraining datasets in Table~\ref{pretraining_dataset}. The most widely used dataset, MIMIC-CXR~\cite{johnson2019mimic} contains chest X-ray images paired with detailed clinical reports. Other datasets like ROCO~\cite{pelka2018radiology} and MediCaT~\cite{subramanian2020medicat} provide a wider variety of image modalities, including radiology and histology domains, and are extracted from a large collection of Open Access articles. Thus, the accompanying text extracted from figure captions tends to be structurally different from clinical reports. Recent efforts on large-scale paired data curation such as Openpath~\cite{huang2023visual} and Quilt-1M~\cite{ikezogwo2023quilt} have seen the utilization of other internet sources such as Twitter and Youtube. These datasets are primarily 2D datasets and contain only 2D slices of the volumetric scans. Meanwhile, the SSPH dataset~\cite{lei2023unibrain} and OASIS~\cite{marcus2007open} have been used to pretrain models on paired 3D MRI scans and reports. 

\subsection{Downstream tasks datasets}
 A large number of datasets have been used for evaluating VLP methods on downstream tasks. We observe an abundance of Chest X-ray datasets being used for the different evaluation tasks, relating to the common use of MIMIC-CXR~\cite{johnson2019mimic} for pretraining.

 \begin{table*}[!ht]
    \centering
    \renewcommand{\arraystretch}{1.2}
    \begin{tabular}{l|l|l|l|l} 
        \hline
        Dataset & Data Modalities & Dataset Size & Classification & Label Distribution \\
        \hline
        
        MELINDA~\cite{wu2021melinda} & Biomedical images, Caption & 5,371 figure-sub-captions& Image/Text & Coarse \& fine-grained classes \\
        
        RSNA Pneumonia~\cite{shih2019augmenting}& Chest X-ray & 30,000 images~(PA/AP views) & Image & 3 classes  \\
        
        CheXpert~\cite{irvin2019chexpert} & Chest-Xray, Reports \footnotemark[1] & 224,316 images~(65240 patients) & Image &  14 multi-label classes\\
        
        CovidX~\cite{wang2003covid} & Chest X-ray & 13975 images  & Image & 3  classes  \\
        
        PatchCamelyon~\cite{veeling2018rotation} & Histopathology~(Lymph node) & 327,680 images & Image & Binary classes  \\
        
        NIH ChestX-ray8~\cite{wang2017chestx} & Chest X-ray & 108,948 images~(32,717 patients) & Image & 8 multi-label classes \\
   
        ChestX-ray14~\cite{wang2017chestx} & Chest X-ray & 112,120 images~(30,805 patients) & Image & 15 multi-label classes  \\
   
        PadChest~\cite{bustos2020padchest} & Chest X-ray, Reports \footnotemark[1] & 160,868 images(69,882 patients)  & Image & 174 findings \& 19 diagnoses  \\
   
        MedMNISTv2~\cite{yang2023medmnist} & Radiology, Histology \& Others& 708,069 2D \& 9,998 3D images & Image & Different in 12 datasets \\
   
        RadNLI~\cite{miura2020improving} & Reports & 960 hypothesis-premise pairs & Text & 3 classes  \\

        MURA~\cite{rajpurkar2017mura} & Bone X-ray & 40,561 images & Image & Binary classes  \\
   
        VinDr-CXR~\cite{nguyen2012vindr} & Chest X-ray & 18,000 images & Image & 28 classes  \\
   
        Shenzhen TB25~\cite{jaeger2014two} & Chest X-ray& Clinical report with 662 images & Image & Binary classes~(TB/Normal)\\
   
        CovidX-CXR 2~\cite{pavlova2022covid} & Chest X-ray & 19,203 images & Image & Binary classes\\
   
        CL25000~\cite{borkowski2019lung} & Histopathology(Lung, Colon) & 25,000 images & Image & 5 cancer classes \\
   
        ChestX-Det10~\cite{liu2020chestx} & Chest X-ray & 3,543 images & Image & 10 classes  \\
   
        MHIST~\cite{wei2021petri} & Histopathology~(Polyps) & 3,152 images & Image & Binary classes~(HP/SSA) \\
   
        Edema Severity~\cite{horng2021deep} & Chest X-ray, Report& 3354 images~(3028 Reports) & Image & 4 classes  \\
   
        MS-CXR-T~\cite{bannur2023learning} & Chest X-ray, Report & 1,326 images & Image & 5 findings, 3 classes \\

        SIIM-ACR PTX& Chest X-ray & 12,954 images~\cite{cheng2023prior} & Image & 2 classes \\

        AIBl \cite{ellis2009australian} & T1 MRI & 1,002 volumetric MRI & Image & 3 classes \\ 

        MIRIAD \cite{malone2013miriad} & T1 MRI & 708 volumetric MRI & Image & 2 classes \\ 
        \hline
    \end{tabular}
    \caption{Classification Datasets}
    \label{tab:classification_datasets}
\end{table*}

\begin{table*}[!ht]
    \renewcommand{\arraystretch}{1.2}
    \begin{tabular}{l|l|l|l|l}
        \hline        

        Task & Dataset &  Data Modalities & Dataset Size  & Remarks/Annotation Type \\
        \hline
        Report& COV-CTR ~\cite{li2023auxiliary} & Lung CT, Report~(Chinese) & 728 image-report pairs & COVID/non-COVID CT scans from \\

        Generation&&&& COVID-CT dataset\\
        
        \hline
        
        &Object CXR  
        \footnotemark[2]
        & Chest X-ray & 10000 images & Bbbox, Ellipse or polygons \\
        
        &LUNA16~\cite{setio2017validation} & Lung CT Scan & 888 images~(1186 annotations) & Location of module with it's size \\
        
        &RSNA Pneumonia~\cite{shih2019augmenting} & Frontal Chest X-Ray & 30000 images & Bounding Box  \\
        
        Object &ISIC 2016~\cite{gutman2016skin} & Dermoscopic Images & 1279 images & Binary Mask \\
        
        Detection&BCCD 
        \footnotemark[3]
        & Blood cell photos  & 364 images & Bounding boxes \\

        &ChestX-Det10~\cite{liu2020chestx} & Chest X-Ray & 3543 images  & Bounding boxes\\
        
        &MS-CXR~\cite{boecking2022making} & Chest X-ray, Texts & 1047 images~(1153 bbox-text pairs) &  Bounding box based on text phase \\

        \hline

        &LUNA16~\cite{setio2017validation} & Lung CT-Scan & 888 images & Binary Mask \\
        
        &SIIM-ACR PTX & Chest Radiographs & 12047 images & Masks \\
        
        Image  &COVID Rural~\cite{desai2020chest, tang2020deep} & Chest X-ray & 221 X-rays(105 patients) & Masks\\
        
        Segmentation&RSNA Pneumonia~\cite{shih2019augmenting} & Chest X-ray & 30000 images~(PA/AP views) & Bounding box  converted into masks \\
        
        &Cremi Competition 
        \footnotemark[4]
        & Brain Electron Micrograph & 3 datasets~(125 3D images each) & Visible neurons on the scans \\

        & BraTS2019 \cite{crimi2016brainlesion} & 4 MRI modalities \footnotemark[5] &  332 volumes &  \\ 
        \hline

        & CheXpert5x200~\cite{huang2021gloria} & Chest X-ray, Report & 200 images per category  & Images from CheXpert competition \\
        
        Retrieval&CheXpert8x200~\cite{zhang2022contrastive} & Chest X-ray, Report & 200 images, 5 sentences per category & Annotated by Radiologist\\
        \hline

        &VQA-RAD~\cite{lau2018dataset} & Radiology & 3,515 QA pairs/315 images & Manually created by clinicians \\
        
        & PathVQA~\cite{he2020pathvqa} & Pathology & 32,799 QA pairs/4,998 images & Obtained from textbook captions   \\
        
        VQA&SLAKE~\cite{liu2021slake} & Radiology & 14,028 QA pairs/642 images & Manually created by clinicians \\
        
        &VQAMed2019~\cite{ben2019vqa} & Radiology & 15,292 QA pairs/3,200 images & From images annotations and QA \\

        &&&& patterns \\
        \hline

    \end{tabular}
    \caption{Other downstream tasks datasets}
    \label{tab:other datasets}
    
\end{table*}

 For image classification, the most commonly used datasets are RSNA Pneumonia~\cite{shih2019augmenting}, CheXpert~\cite{irvin2019chexpert}, CheX-ray8(NIH X-ray)~\cite{wang2017chestx} and contain Chest X-rays. MELINDA~\cite{wu2021melinda} offers both biomedical images and caption pairs for image, text as well and multimodal evaluation but the imaging modality contains graphs and plots besides medical scans and is vastly different from the pretraining datasets available. MedMNIST~\cite{yang2023medmnist} offers a wide variety of datasets with both 2D and 3D images spanning Radiology, Histology, and other scans but the images are resized to small dimensions(28x28 or 28x28x28) and may not be suitable for zero-shot evaluation as pretraining is done at higher resolution level. MS-CXR-T offers images with progression stage labels and can be used for temporal image classification We refer the readers to Table~\ref{tab:classification_datasets} for a list of the commonly used classification datasets.

 The most widely used VQA datasets include VQA-RAD~\cite{lau2018dataset}, PathVQA~\cite{he2020pathvqa}, SLAKE~\cite{liu2021slake} and VQAMed2019~\cite{ben2019vqa}. Besides PathVQA, which contains histopathology images, other datasets contain 2D radiology scans such as X-rays and single slice CT and MRI. These datasets contain both close-ended questions which have a fixed set of possible answers and open-ended ones with a large variety of possible answers. SLAKE also contains open-ended knowledge-based questions that require the model to have a broader domain knowledge besides requiring a proper understanding of the image.
 For report generation, most works have opted to use the paired image text datasets used for pretraining as the task requires both images and text descriptions during training. 
 We have omitted these entries in Table~\ref{tab:other datasets} and refer the readers to Table~\ref{pretraining_dataset} for a list of paired datasets.

 Object-CXR, RSNA Pneumonia~\cite{shih2019augmenting} and NIH Chest X-ray~\cite{wang2017chestx} are frequently used for evaluation on object detection. Object-CXR contains annotations for detecting foreign objects in Chest X-rays whereas RSNA Pneumonia and NIH Chest X-ray contain bounding box annotations identifying pathological findings in Chet X-rays. Other datasets like LUNA16~\cite{setio2017validation} for nodule detection from lung CT scans, Lesion Detection Dataset and ISIC 2016~\cite{gutman2016skin} for skin lesion detection on dermatological scans have also been used in some of the works. Another dataset MS-CXR~\cite{boecking2022making} provides bounding box annotations based on accompanying text descriptions on a subset of Chest X-rays from MIMIC-CXR.

 For image segmentation, the most commonly used datasets include SIIM-ACR Pneumothorax, COVID Rural~\cite{desai2020chest, tang2020deep} and RSNA Pneumonia~\cite{shih2019augmenting}, all involving masks corresponding to pathological findings in Chest X-rays. Dataset from Cremi Competition providing masks for neurons on 3D Electron Microscopy Scans of Adult~\textit{Drosophila Melanogaster}, and BraTS2019~\cite{crimi2016brainlesion} containing 3D MRI scans have been used to evaluate on 3D segmentation.

 For retrieval, many works opt for the use of a subset of paired image text datasets like MIMIC-CXR and ROCO. Another dataset CheXpert8x200~\cite{zhang2022contrastive} was first introduced by ConVIRT which allows for both image-image and text-image retrieval by collecting 200 candidate images with 5 text queries for each of 8 disease categories in CheXpert and curating 10 image or text queries with the help of radiologists. Many works have adapted this approach to curate their own version, such as the CheXpert5x200~\cite{huang2021gloria} by GLORIA which instead used reports available within CheXpert and performed image-to-text retrieval, or CheXpert5x200 by PRIOR, LIMITR and MedCLIP which instead used sampled MIMIC-CXR reports due to the unavailability of CheXpert reports. Similarly, some works have also used MIMIC5x200 using both images and reports from MIMIC-CXR.

\footnotetext[1]{Reports publicly not available}

\section{Limitation and Challenges in Medical VLP}
\label{limitations}
Despite the effectiveness of Medical VLP across different tasks and scenarios, Medical VLP faces different limitations and challenges. We discuss these outlining existing and potential solutions in this section.

\subsection{Pretraining Data size and coverage}
A significant challenge in the medical domain is the scarcity of large-scale paired image-text data. Paired datasets undergo extensive de-identification processes and often necessitate patient permission before being made publicly available. This makes it challenging to curate even small-scale datasets from health centers. For instance, the widely used pretraining dataset, MIMIC-CXR~\cite{johnson2019mimic}, contains only a few hundred thousand images and text reports. Moreover, most medical image-text datasets primarily focus on Chest X-rays and neglect other types of medical scans and anatomical locations. 

\begin{figure*}[h!]
\centering
\begin{subfigure}[t]{0.41\textwidth}
    \includegraphics[width=\linewidth]{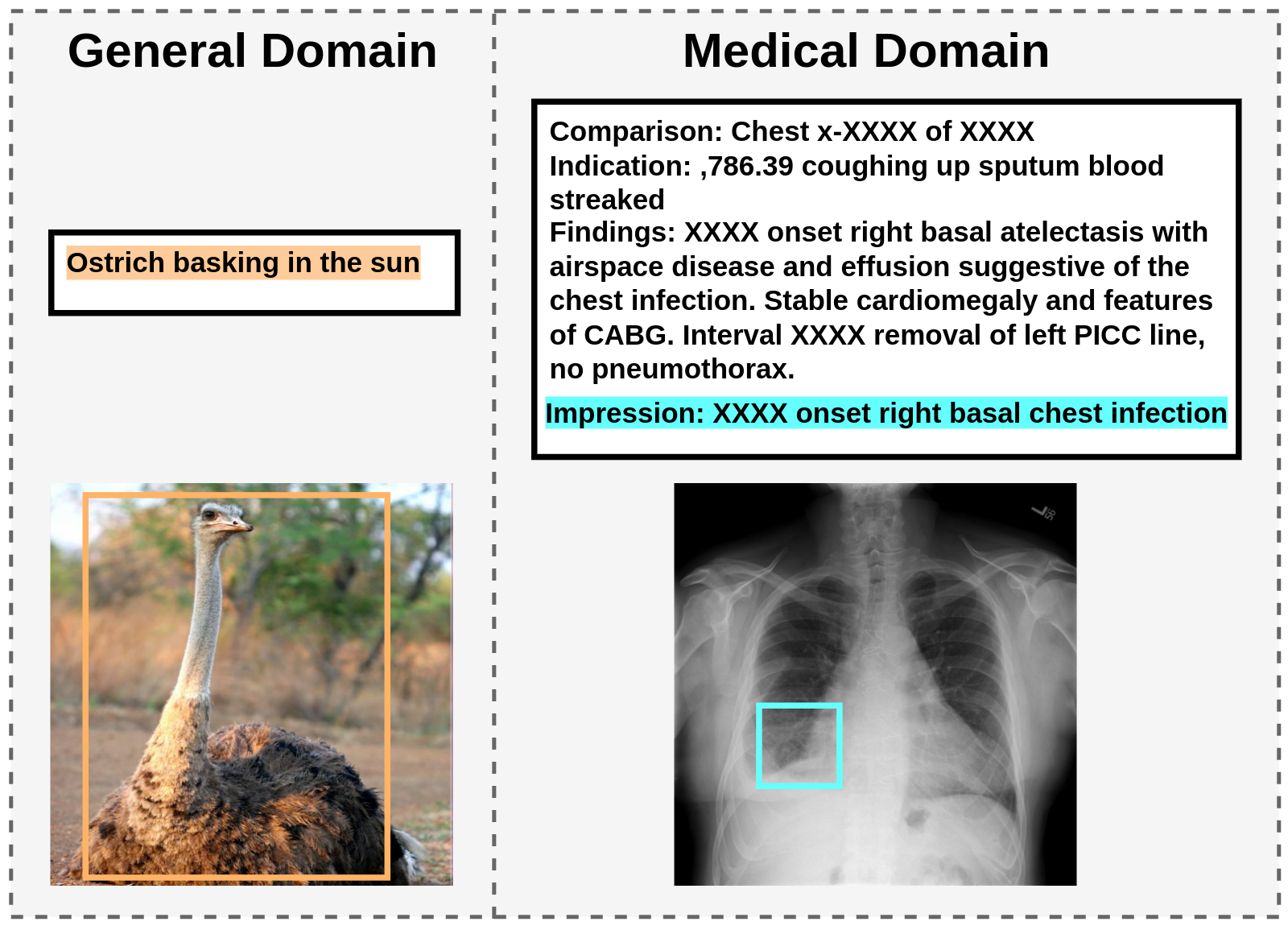}
    \caption{Illustration of domain-specific nuances in medical dataset, samples from LAION-400M~\protect\cite{schuhmann2021laion}~(left) and Open-I IU X-ray collection~\protect\cite{demner2016preparing}~(right)}
    \label{fig:nuance}
\end{subfigure}
\begin{subfigure}[t]{0.41\textwidth}
    \includegraphics[width=\linewidth]{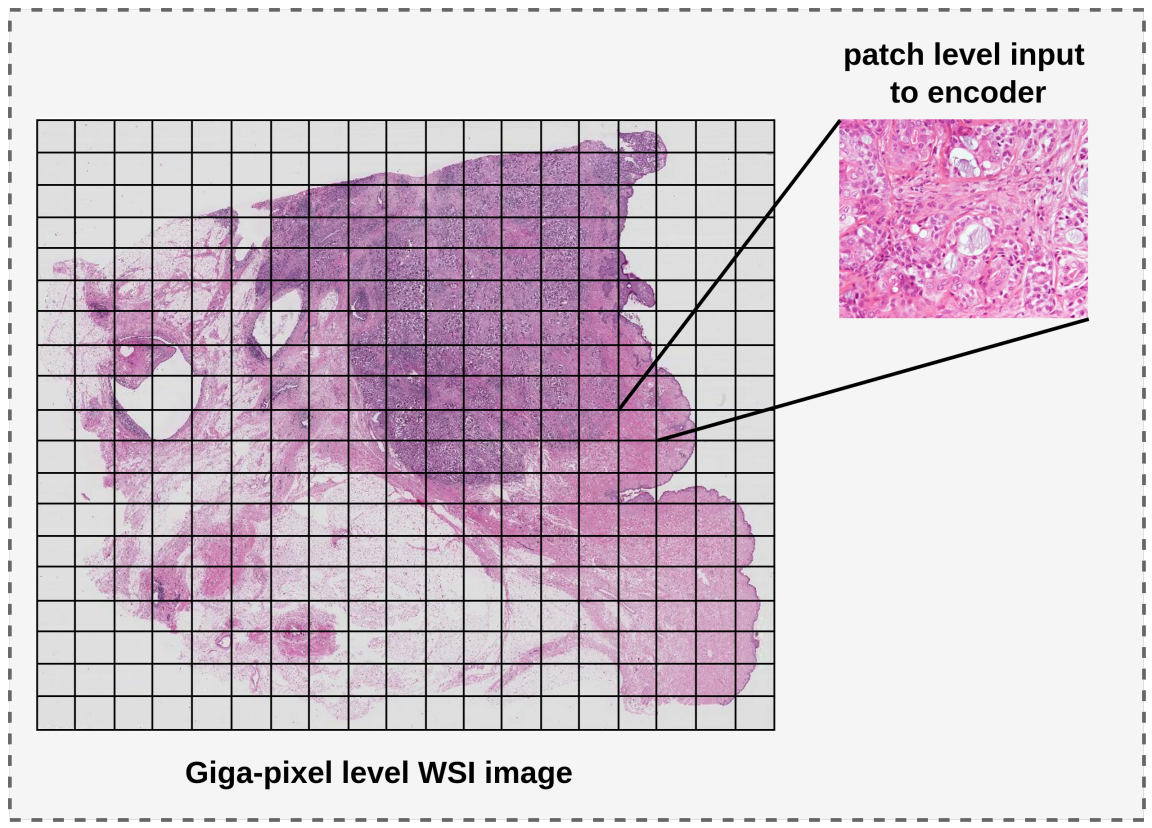}
    \caption{WSI level pathology image}
    \label{fig:pathology}
\end{subfigure}
\begin{subfigure}[t]{0.248\textwidth}
    \includegraphics[width=\linewidth]{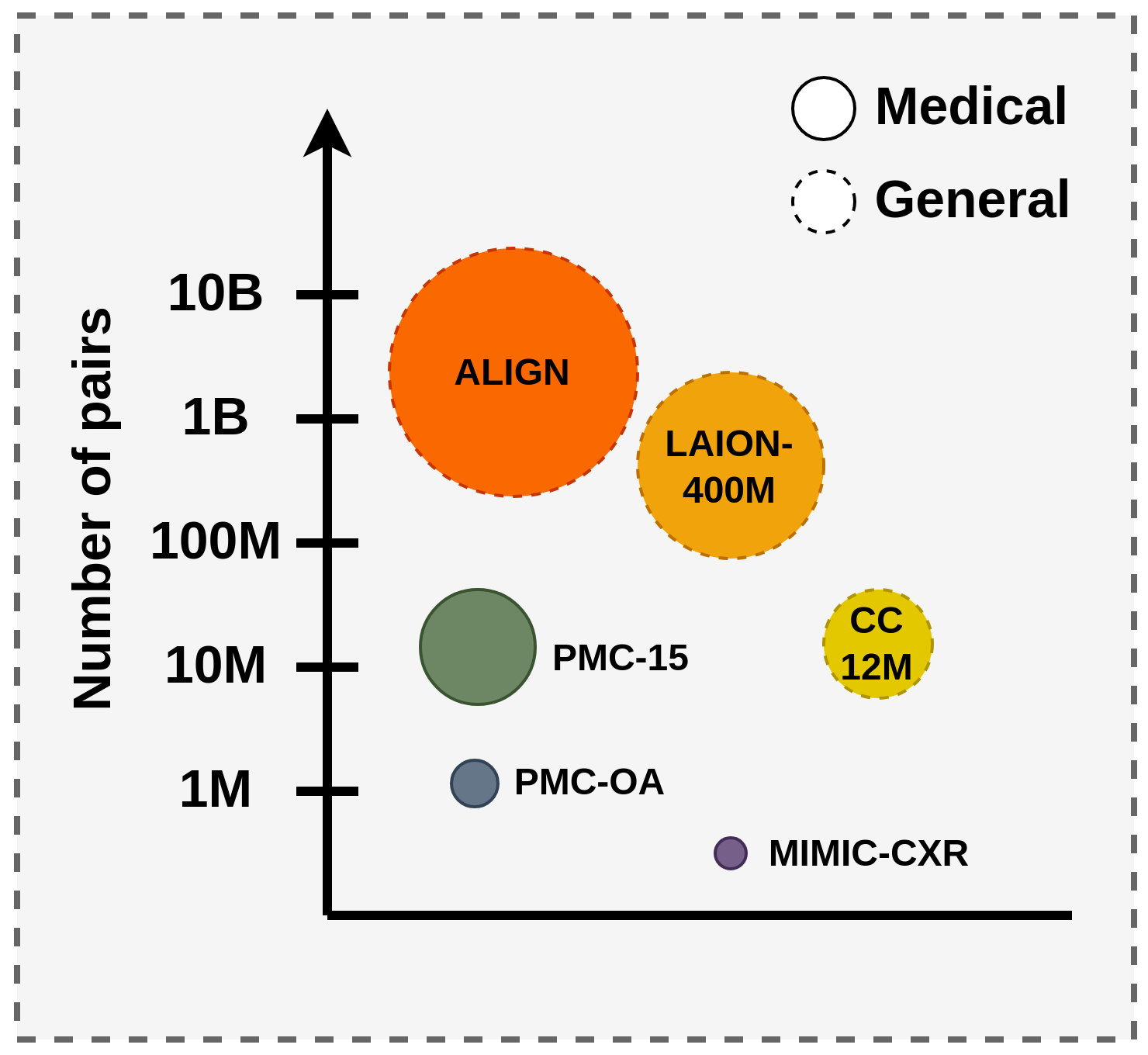}
    \caption{Difference in scales of pretraining datasets across domains}
    \label{fig:scale}
\end{subfigure}
\begin{subfigure}[t]{0.275\textwidth}
    \includegraphics[width=\linewidth]{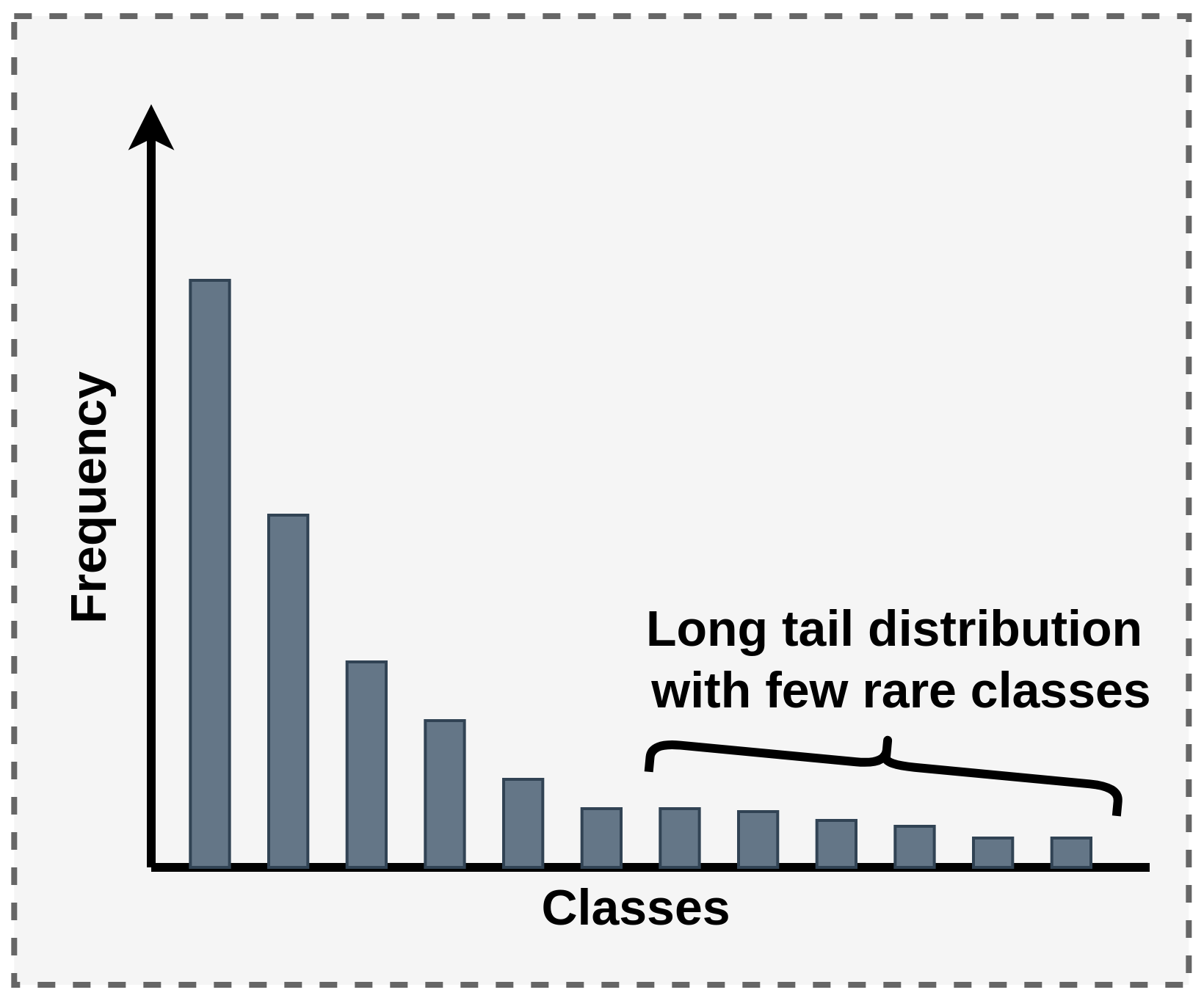}
    \caption{Long tail distribution common in medical datasets}
    \label{fig:long_tail}
\end{subfigure}
\begin{subfigure}[t]{0.30\textwidth}
    \includegraphics[width=\linewidth]{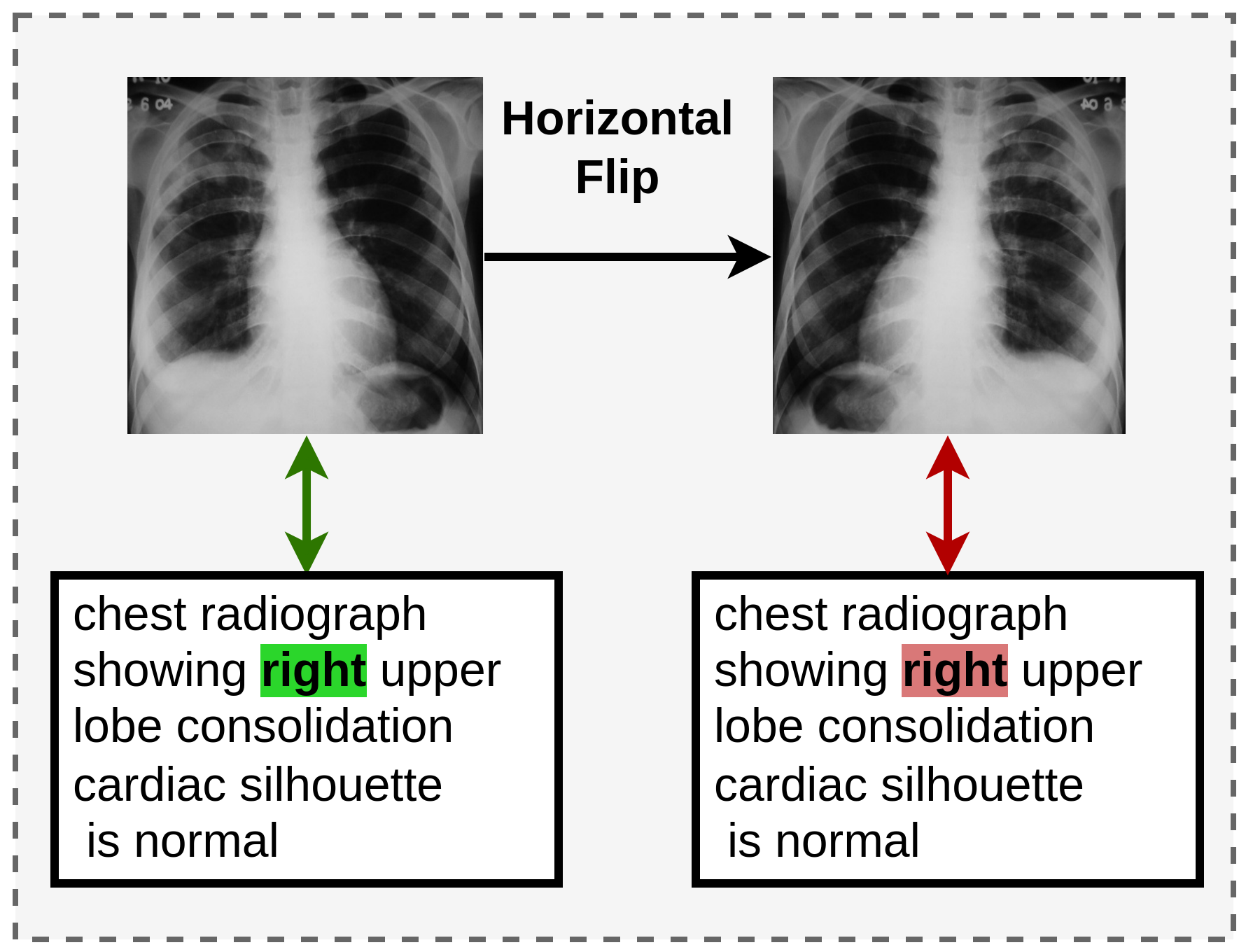}
    \caption{Image augmentations causing potential misalignment with text}
    \label{fig:augmentation}
\end{subfigure}
\caption{Limitations and Challenges in Medical VLP} 
\label{fig:challenges}
\end{figure*}

Recently, researchers have made efforts to compile larger datasets by extracting medical images and text from PubMed articles, offering more variety~\cite{zhang2023large, lin2023pmc}. Other works propose employing internet sources such as Twitter and YouTube videos to compile large-scale paired datasets~\cite{ikezogwo2023quilt, yuan2023learning}. Nevertheless, these datasets still fall far short of the extensive datasets utilized in general domain vision-language pretraining, where paired data scales can range from hundreds of millions~\cite{radford2021learning} to even billions of pairs~\cite{jia2021scaling} as illustrated in Figure~\ref{fig:scale}.

\footnotetext[2]{\url{https://web.archive.org/web/20201127235812/https://jfhealthcare.github.io/object-CXR/}}
\footnotetext[3]{\url{https://github.com/Shenggan/BCCD_Dataset}}

Federated learning has also emerged as a promising solution to address the challenges of data privacy that are limiting the data curation process. It involves a decentralized process where instead of collecting data from different sources for model training, a client model is instantiated at each data location to perform training locally. Then the separate learnings are combined to improve a central server model.
A notable work, FedMedVLP~\cite{lu2023scaling} applied federated learning specifically to medical VLP where different datasets are associated with separate client models. This approach offers several potential advantages. Firstly, it eliminates the need for direct data sharing between institutions, thereby safeguarding patient privacy. Secondly, it enables the curation of large-scale pretraining data from multiple health centers without compromising data security. Similarly, the use of unpaired data and synthetic data generation can also help alleviate the data scarcity issues.

\subsection{Medical Domain Nuances}
Vision-language pretraining for medical applications presents unique challenges. 
Radiology reports in the medical field often exhibit characteristics distinct from natural text. 
Notably, they frequently contain uncertainty and negation keywords, which are essential for precise medical interpretation~\cite{boecking2022making}. 
Furthermore, medical reports tend to be longer than typical image descriptions. 
In addition, medical images often share visual similarities and only display subtle differences, in the presence of abnormalities~\cite{huang2021gloria}. Consequently, models pretrained on general-domain data tend to underperform when applied to the medical domain. 
While leveraging large-scale pretraining data can be beneficial to capture these nuances, obtaining unimodal data on a large scale is often much more feasible. It becomes imperative to engage in large-scale domain-specific pretraining of the modality encoders to effectively capture these nuances and enhance VLP~\cite{boecking2022making, lu2023visual}.

\subsection{Augmentations}
\footnotetext[4]{\url{https://cremi.org/}}
\footnotetext[5]{T1, T1 contrast-enhanced, T2 and FLAIR}
Augmentations are crucial to improving the generalizability of VLP models, especially with the limited pretraining datasets available.
However, applying data augmentations from the general domain to medical images is not always feasible. For instance, standard augmentations like random image cropping can inadvertently remove crucial regions, especially when abnormalities are localized in medical images~\cite{huang2023self}. And some spatial image augmentations can bring misalignment with the corresponding text description as illustrated in Figure~\ref{fig:augmentation}. 

Similarly in the context of text augmentations, only simple approaches like sentence/section shuffling and sampling are generally used. However, the sampled sentence alone may be ambiguous and can describe multiple images resulting in a weaker text supervision for the accompanying image. 

Careless augmentation can impact the model's ability to generalize effectively. Therefore, a more thoughtful approach to data augmentation is required when working with medical images to avoid compromising the integrity of the images and their alignment with corresponding reports. Another potential solution is the use of multimodal augmentations that take into consideration both the input modalities. For instance, LeMDA~\cite{liu2022learning}, a learnable augmentation module that performs multimodal augmentation in the feature space was seen to be more effective than augmenting each modality separately during supervised training. However, the effectiveness of such an approach when applied on self-supervised objectives may require further study. Meanwhile, for text, utilizing LLMs for augmentation can also be beneficial to improve pretraining~\cite{fan2023improving}.

\subsection{Class Imbalance}
As illustrated in Figure~\ref{fig:long_tail}, medical datasets often exhibit significant class imbalances and long tail data distribution, with the majority of samples representing normal cases devoid of abnormal findings, or very common pathologies. Consequently, models trained on such datasets may perform poorly when faced with rare disease samples. Additionally, the class imbalance introduces a considerable number of false negatives in contrastive learning while pretraining~\cite{boecking2022making}. 
Some methods attempt to mitigate this false negative issue by incorporating weights or supervising the contrastive loss with a predicted similarity index~\cite{wang2022medclip, chen2023knowledge}. However, these approaches frequently rely on additional image labels or external tools, which can have limited coverage and applicability.

A potential solution to mitigate the effect of such class imbalances could be the use of feature compensation techniques such as the expansion of the feature space by considering the features to belong to a distribution. The introduction of a more diverse set of features can help the model generalize better to minority classes. Feature expansion can be applied in conjunction with techniques like coreset sampling where a subset of the most representative samples are selected. This process helps to ensure that the model is not biased towards specific features occurring more frequently in the dataset. While these approaches can be beneficial during downstream fine-tuning, applying these techniques in multimodal and self-supervised scenarios demands further study.

\subsection{ Computation Pathology challenges}
The field of computational pathology in medicine presents its unique set of challenges. As illustrated in Figure~\ref{fig:pathology}, Whole Slide Images~(WSI) in pathology encompass gigapixel-level images, while paired image-text pathology datasets usually operate at the patch level derived from these WSIs. Consequently, even adapting a VLP model pretrained on patch-level images becomes challenging when applied to WSI images. Handling an entire WSI demand at once demands substantial memory, and current medical VLP image encoders aren't optimized for this task.

Towards this end, MI-Zero~\cite{lu2023visual} introduced a strategy based on Multi-Instance Learning~(MIL) to adapt VLP models for zero-shot WSI classification by aggregating patch-level predictions. However, this approach, though parallelizable, can still pose practical challenges for large datasets. Storing the entire WSI image in RAM is cumbersome, requiring the individual patches to be read from the disk, which can hinder efficiency.

\subsection{Multiple imaging modalities per study}
Generalizing the scenario in Section~\ref{different_views}, clinical investigations often involve multiple types of radiological scans. MRI alone encompasses four distinct modalities, while retinal imaging utilizes methods like fundus imaging and Optical Coherence Tomography, among others. Different modalities excel in identifying specific abnormalities, prompting clinicians to employ various scans for accurate assessments. Constructing a model capable of working across these diverse imaging modalities of a single study effectively might necessitate using multiple modalities during pretraining. However, obtaining such paired data for pretraining may be challenging. Moreover, when utilizing multiple modalities during pretraining, the model should be capable of handling missing modality inputs as not all datasets or tasks may encompass all modalities, requiring architectural decisions to address these potential gaps.

Towards this end, UniBrain~\cite{lei2023unibrain} curated and pretrained a VLP model incorporating the four MRI modalities. The accompanying report was decomposed into modality-specific texts and aligned with the corresponding image modality and the aggregated features from all images were aligned with the global report. However, UniBrain still struggled with missing modality scenarios.

\section{Perspectives and Future Directions}
\label{future directions}
Considering privacy concerns among clients regarding sharing datasets on a large scale, federated learning can emerge as a potential direction to facilitate large-scale VLP through decentralized client-level training. FedMedVLP~\cite{lu2023scaling} introduced a federated learning framework for medical VLP to address privacy concerns and scalability by utilizing separate client-associated datasets. However, they still overlook challenges inherent in federated learning itself, such as client dropout, communication overhead, and security threats. Addressing these issues is crucial to leveraging private health center data without costly curation.

Similarly, the substantial computational resources needed for the training and deployment of medical VLP pose a significant challenge, especially in limited hardware settings. Recent advancements, including knowledge distillation, parameter pruning, and quantization~\cite{xu2023survey}, can be employed to enhance the accessibility of VLP models. This application of techniques enables the deployment of VLP models on edge devices like smartphones, facilitating early diagnosis and extending their reach to remote areas.

On another note, while many studies focus on chest X-rays, very few explore 3D imaging inputs and raw surgical videos, possibly due to a lack of suitable pretraining and benchmark datasets. Curating paired pretraining and proper benchmarking datasets across diverse medical image modalities is vital for the holistic advancement of medical VLP.

We've also observed a recent surge of interest in the creation of large-scale unified foundation models in medical VLP \cite{wu2023towards, zhang2023biomedgpt, li2023llava}. Although RadFM~\cite{wu2023towards} adeptly handles both 2D and 3D inputs, it falls short in generating image annotations or handling video inputs. The next crucial step in advancing comprehensive medical AI involves the development of a unified model accommodating all healthcare modalities, encompassing genomic data, Electronic Health Records (EHR), scans, reports, and time series data~\cite{moor2023foundation}. Attaining this goal necessitates not only extensive data but also advancements in model architectures and training techniques.

While this review focuses on several technical aspects, the clinical implications of medical VLP models—specifically their interpretability and integration into healthcare settings—remain largely unaddressed. 
Future work should delve into these critical aspects to bridge the gap between technical advancements and practical healthcare implementation. 
Additionally, the limited time frame and search strategy during our literature review may have resulted in the omission of several pertinent works in this field. Including a more comprehensive array of medical VLP works should be a priority in future iterations of this survey.

\section{Conclusion}
Medical VLP holds significant promise in addressing the scarcity of labeled data within the medical domain, while also providing a valuable advantage by supporting various downstream tasks through pretraining. This survey extensively investigated the landscape of medical VLP, examining its various aspects, including pretraining objectives, architecture, downstream evaluation tasks, and datasets. Additionally, we delved into several challenges within medical VLP, encompassing issues related to pretraining data, augmentation techniques, domain-specific nuances, computational pathology, and the integration of multiple imaging modalities. Finally, we have presented our perspectives and outlined several future directions in the field. We believe that this survey will serve as a valuable resource for both researchers and clinicians interested in the advancement of medical AI methodologies.

\bibliographystyle{IEEEtran}
\bibliography{references}
\vspace{12pt}

\end{document}